\documentclass{article}

 \usepackage[preprint]{neurips_2026}

\usepackage[utf8]{inputenc} 
\usepackage[T1]{fontenc}    
\usepackage{hyperref}       
\usepackage{url}            
\usepackage{booktabs}       
\usepackage{amsfonts}       
\usepackage{nicefrac}       
\usepackage{microtype}      
\usepackage{xcolor}         

\usepackage{graphicx}
\usepackage{subfigure}
\usepackage{adjustbox}
\usepackage{multirow}
\usepackage{amsmath, amssymb}
\usepackage{wrapfig}
\usepackage[table]{xcolor}
\usepackage[most]{tcolorbox}
\usepackage{caption}
\usepackage{ulem}
\usepackage{algorithm}
\usepackage{algorithmic}

\title{Planning with Unified Multimodal Models}

\author{
  Yihao Sun\textsuperscript{1,2}\thanks{Correspondence to: Yihao Sun <\texttt{yihao.sun@mila.quebec}>}
 \quad
  Zhilong Zhang\textsuperscript{3} \quad
  Yang Yu\textsuperscript{3} \quad
  Pierre-Luc Bacon\textsuperscript{1,2} \\\\
  \textsuperscript{1} Mila - Qu\'ebec AI Institute \quad
  \textsuperscript{2} Universit\'e de Montr\'eal  \quad
  \textsuperscript{3} Nanjing University \quad 
}

\begin{document}

\maketitle

\begin{center}
    \vspace{-7mm}
    \includegraphics[width=0.95\textwidth]{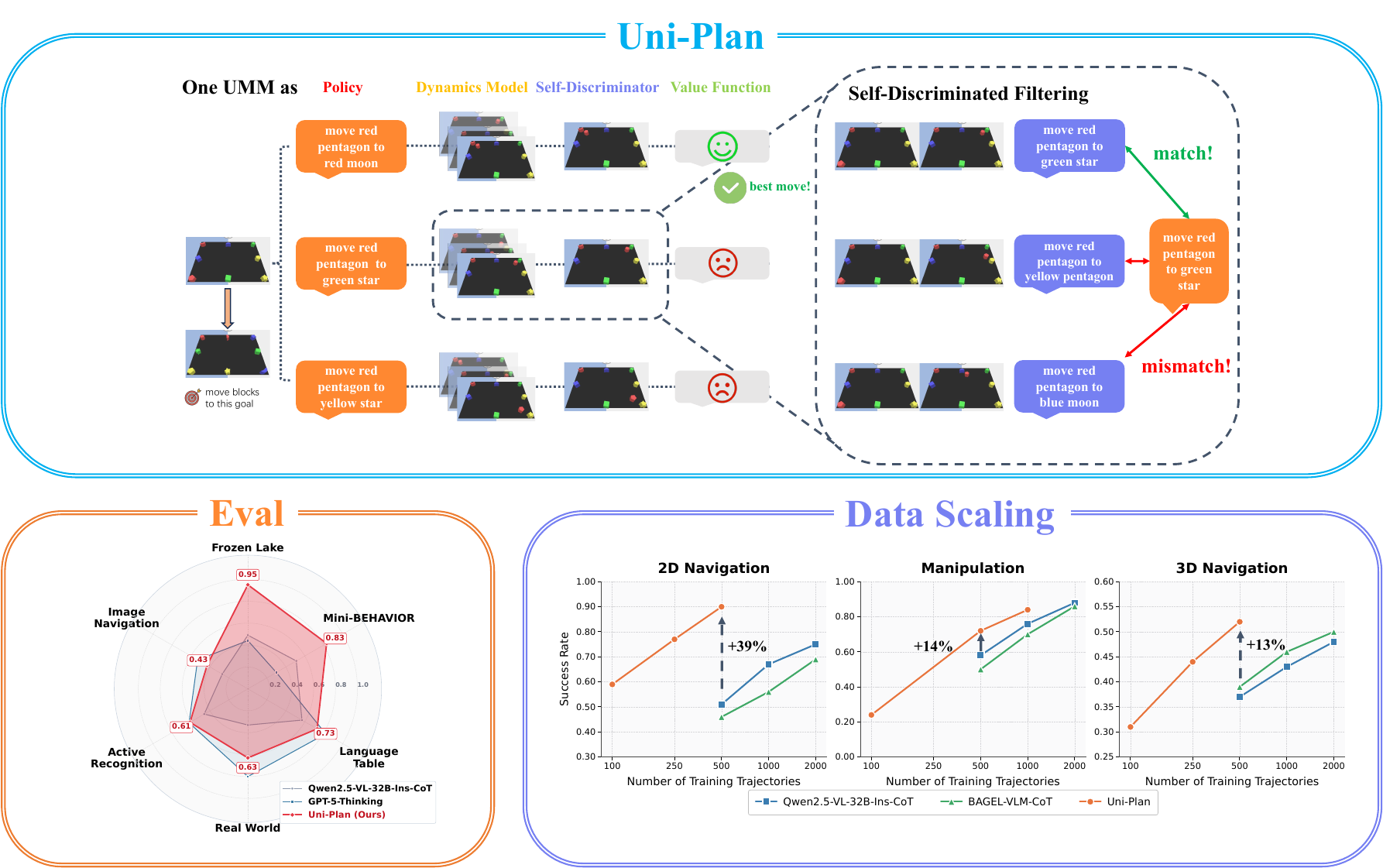}
    \captionof{figure}{Overview of Uni-Plan, a planning system where one single UMM serves as (i) the policy,
(ii) the dynamics model, (iii) the self-discriminator, and (iv) the value function simultaneously. Uni-Plan outperforms VLMs in multiple decision-making scenarios and also shows better data scaling trend.}
    \label{fig:overview}
\end{center}

\begin{abstract}
With the powerful reasoning capabilities of large language models (LLMs) and vision-language models (VLMs), many recent works have explored using them for decision-making. However, most of these approaches rely solely on language-based reasoning, which limits their ability to reason and make informed decisions. Recently, a promising new direction has emerged with unified multimodal models (UMMs), which support both multimodal inputs and outputs. We believe such models have greater potential for decision-making by enabling reasoning through generated visual content. To this end, we propose \textit{Uni-Plan}, a planning framework built on UMMs. Within this framework, a single model simultaneously serves as the policy, dynamics model, and value function. In addition, to avoid hallucinations in dynamics predictions, we present a novel approach \textit{self-discriminated filtering}, where the generative model serves as a self-discriminator to filter out invalid dynamics predictions. Experiments on embodied decision-making tasks show that Uni-Plan substantially improves success rates compared to VLM-based methods, while also showing strong data scalability, requiring no expert demonstrations and achieving better performance under the same training-data size. This work lays a foundation for future research in reasoning and decision-making with UMMs.\looseness=-1
\end{abstract}
\section{Introduction}
Large language models (LLMs) and vision-language models (VLMs) have demonstrated strong reasoning capabilities across a wide range of tasks~\citep{llm-few-shot-learner, cot, gpt-4, mcot}. Motivated by this, many recent works~\citep{llm-zero-shot-planner, saycan, palm-e, look-before-you-leap} have explored their application to decision-making, such as generating high-level, step-by-step plans for long-horizon tasks. However, their planning process remains purely text-based. Even for most VLMs, visual inputs are typically used only at the initial stage of reasoning, rather than being incorporated throughout the thinking process. As a result, the reliance on a single modality limits the model’s ability to accurately represent the current state during planning. The absence of multimodality throughout the thinking process limits their effectiveness in complex scenarios that require accurate spatial or visual understanding.

Recently, an increasing number of works~\citep{visual-sketchpad, image-of-thought, mvot, thinking-with-images} have proposed incorporating images into the reasoning process. This is typically achieved by integrating external tools to interpret visual observations, for example, by generating program code to call Python plotting libraries~\citep{visual-sketchpad}, or by invoking vision models for segmentation or object detection on input images~\citep{image-of-thought}. However, such approaches are highly dependent on separate visual modules or external toolchains, which limits their adaptability to more complex visual reasoning tasks. In contrast, a different line of work~\citep{mvot, thinking-with-images, uni-cot, thinkmorph} explores generating intermediate images directly within the model to support reasoning. Although this approach is more general and holds greater promise, it has been used primarily to visualize reasoning traces~\citep{mvot} or assist text reasoning~\citep{zebra-cot,thinkmorph}/image generation~\citep{thinking-with-images,uni-cot}, rather than to enable more sophisticated decision-making.

Notably, a promising new class of models has recently emerged, i.e., unified multimodal models (UMMs)~\citep{next-gpt, showo2, emu3, mogao, bagel}, which support both multimodal inputs and outputs, typically in the form of images and text. We argue that such models are particularly well-suited for decision-making, as they can simultaneously serve as dynamics models (generating the next visual observation), as policies (generating text-based actions), and as value functions (estimating which move is the best), thus enabling integrated planning. However, such models remain subject to the curse of the horizon, with a key bottleneck being their limited ability to serve as faithful dynamics models. While current state-of-the-art models can perform basic image-editing tasks~\citep{bagel, mogao}, our findings indicate that they are still insufficiently accurate to give reliable dynamics predictions. This limitation can be alleviated through finetuning for relatively simple downstream tasks, but the improvement does not extend to more challenging scenarios, particularly those involving stochastic dynamics.

To address this challenge, we strategically leverage the UMM's flexible input–output modality to employ it as a self-discriminator for filtering out invalid transition predictions. Concretely, the model first generates multiple candidate predictions for the next observation. It then operates in an inverse dynamics mode, inferring the action that would lead from each current–next observation pair. By comparing these inferred actions with the actual action, we can identify and discard those transitions where the actions do not match, effectively removing implausible predictions. Building on this capability, we develop a planning framework \textit{Uni-Plan}, and demonstrate its superior performance across a range of embodied decision-making tasks.

We highlight the main contributions of our work below:
\begin{itemize}
    \item We propose \textit{self-discriminated filtering}, where the generative model serves as a self-discriminator to filter out invalid dynamics predictions for a more accurate dynamics model.
    \item We present a planning framework \textit{Uni-Plan}, illustrated in Figure~\ref{fig:overview}, where one UMM plays the roles of (i) policy, (ii) dynamics model, (iii) self-discriminator, and (iv) value function simultaneously.
    \item Evaluating on 6 embodied decision-making tasks, our method achieves nearly 30\% higher success rates than open-source VLM-based planning methods, and even matches the powerful GPT-5-Thinking model. Furthermore, our method also exhibits strong data scalability, requiring no expert demonstrations for finetuning and outperforming VLMs when trained with the same amount of data.
\end{itemize}
\section{Planning with Unified Multimodal Models}

\subsection{Formulation}
In this work, we focus on leveraging unified multimodal models (UMMs) for decision-making. Here, the UMMs refer to such models capable of multimodal inputs and outputs, typically in the form of images and text. This kind of model can give us more flexibility and higher potential when using them for decision-making. In this paper, we mainly use BAGEL~\citep{bagel}, the state-of-the-art open-source UMM, as the foundation model. We refer readers to Appendix~\ref{appendix:intro_to_bagel} for more details on BAGEL, and to Appendix~\ref{appendix:other_umms} for results with other UMMs.

We formulate the decision-making process as a hierarchical framework. At the high level, given an initial visual observation $o_0$ of the environment and a goal $g$ described in natural language, A model (VLM/UMM) is required to generate a sequence of plans $a_{0:H}$ to achieve the goal, where each $a_i$ is a textual action. At the low level, we assume the availability of a set of off-the-shelf policies (skills) $\{\pi^i\}_{1:N}$ that serve as controllers, producing low-level control actions conditioned on the current observation and the corresponding textual action. These low-level policies can be derived from either behavior cloning or reinforcement learning. Our work primarily focuses on the high-level planning component.

The core idea of our method is to employ a UMM for planning, which inherently integrates the functions of a dynamics model, a policy, and a value function. Owing to the flexible input-output modalities of UMMs, a single model can simultaneously serve all these roles. In Section~\ref{2.2}, we first describe how to utilize it as a reliable dynamics model, which constitutes the most challenging aspect of the planning system. Subsequently, in Section~\ref{2.3}, we present the design of the overall planning framework.

\subsection{UMMs as dynamics models}
\label{2.2}
To enable high-level planning, the model must be capable of predicting the next visual observation conditioned on the current observation and a textual action, i.e., $P_{\mathrm{UMM}}(o_h, a_h) \rightarrow o_{h+1}$. This task closely resembles image editing in the training of UMMs~\citep{bagel, mogao}, as both require accurate language grounding while preserving the consistency of unaffected regions in the image. However, dynamics prediction presents a greater challenge, as it additionally demands the ability to reason about precise causal effects of textual actions.

\begin{figure*}[!t]
    \centering
    \vspace{-12mm}
    \includegraphics[width=0.9\linewidth]{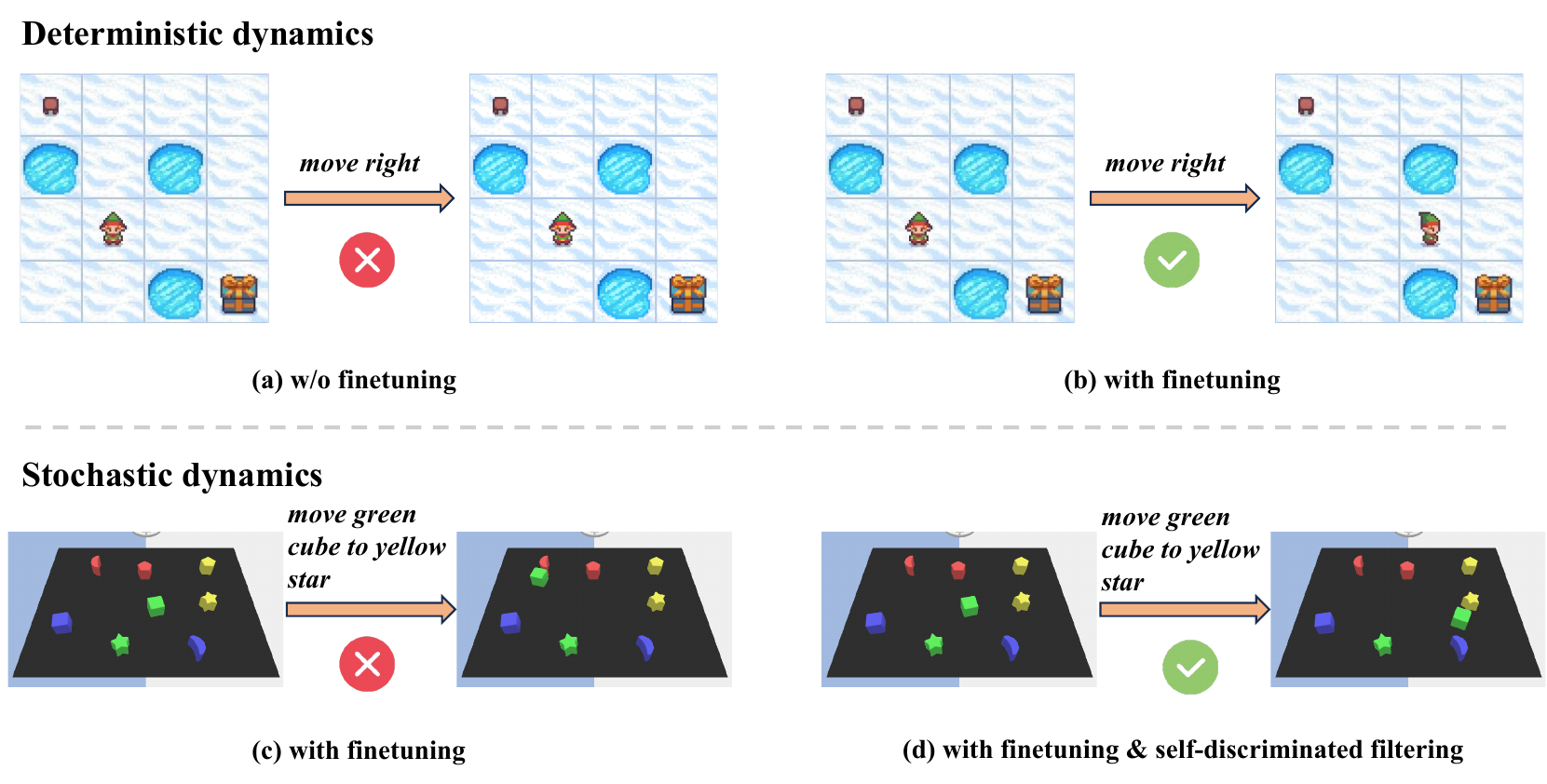}
    \vspace{-2mm}
    \caption{Illustration of dynamics predictions by different models.}
    \label{fig:dynamics_illustration}
    \vspace{-5mm}
\end{figure*}

As an illustrative example, we employ BAGEL~\citep{bagel} to perform dynamics predictions in a maze environment (Figure~\ref{fig:dynamics_illustration}(a)). Note that we directly use its open-source weights without any finetuning. While the model preserves overall image consistency, it fails to predict the character’s position accurately. This limitation can, however, be mitigated through few-shot finetuning. As shown in Figure~\ref{fig:dynamics_illustration}(b), after such adaptation, the model can serve as an effective dynamics model.

However, we find that only finetuning remains insufficient for accurate dynamics predictions on more challenging tasks, particularly those involving stochastic dynamics. In Figure~\ref{fig:dynamics_illustration}(c), we illustrate the dynamics prediction for table rearrangement tasks using the finetuned BAGEL. The stochasticity in this setting arises from the existence of multiple valid outcomes $\{o^i_{h+1}\}$ for a given $(o_h, a_h)$. For example, suppose that the text action is ``\textit{move green cube to yellow star}’’, such that multiple valid next states may exist because the green cube could be placed at different relative positions around the yellow star, leading to different but equally correct results.

To mitigate this issue, we propose \textit{self-discriminated filtering}, which lets the model act as its own discriminator when selecting among sampled dynamics predictions. The key insight is that \textit{large models often exhibit stronger understanding and verification abilities than faithful generation}, a phenomenon also reflected in prior work on self-verification and self-correction~\citep{llm_self_veri,self-check,s2r}. Concretely, we jointly train a UMM for \textit{inverse dynamics inference}, $P^{-1}_{\mathrm{UMM}}(o_h, o_{h+1}) \rightarrow a_h$, and use it to verify candidate predictions $\{\hat{o}_{h+1}^i\}$ generated for $(o_h, a_h)$. A candidate is kept only if the inferred action from $(o_h, \hat{o}_{h+1}^i)$ matches the ground-truth action $a_h$. To further suppress rare failures such as missing or duplicated objects, we additionally apply an object-count consistency check between the current and predicted observations. This filtering substantially reduces hallucinations, as shown in Figure~\ref{fig:dynamics_illustration}(d); quantitative results are provided in Section~\ref{sec:3.2}.

\subsection{Whole Planning System Design}
\label{2.3}
In this section, we describe the complete planning system built upon UMMs. Beyond the dynamics model, the planner additionally requires a mechanism for proposing candidate behaviors (policy) and a criterion for selecting among their imagined consequences (value). We next describe two planner algorithms built on this framework: an open-loop planner tailored to fully observable settings and a closed-loop planner designed for partially observable settings, each paired with a different instantiation of the policy and value function. See corresponding pseudocode in Appendix~\ref{appendix:impl_details}.

\vspace{-3mm}
\paragraph{Open-loop Planning.} When the task is fully observable, a natural design is open-loop planning, in which the planner optimizes a complete action sequence from the initial observation and then executes it without consulting observations. In this setting, we instantiate the policy as a prompted UMM $\pi_{\mathrm{UMM}}(o_t,g)\rightarrow a_t$ that predicts the next textual action given the current imagined observation $o_t$ and the goal $g$. We perform beam search in the imagined state space. Specifically, at each planning step, the policy samples $K$ candidate actions for every active beam, and the dynamics model rolls out each candidate for one step. The resulting successor states are filtered using the self-discriminated filtering mechanism introduced in Section~\ref{2.2} and then scored by a heuristic value function. Following \cite{vlp}, we instantiate this value estimator as a finetuned UMM $H_{\mathrm{UMM}}(o,g) \rightarrow u$ that predicts the remaining number of steps needed to satisfy $g$ from observation $o$. The planner retains the top-$B$ partial trajectories and repeats this expansion process until the maximum planning horizon is reached. This design is simple and effective when the observation fully reveals the underlying state.

\vspace{-3mm}
\paragraph{Closed-loop Planning.} To handle partially observable environments, we adopt a closed-loop planning scheme inspired by the \textit{``proposal, simulation, and revision''} paradigm of World-in-World~\citep{worldinworld}. Here, the policy is instantiated differently: rather than proposing a single next action for beam expansion, it directly generates $K$ candidate action sequences $\{a^{(k)}_{0:H-1}\}_{k=1}^{K}$, each with horizon $H$, conditioned on the current observation and the goal. In the \textit{simulation} stage, the dynamics model rolls out each candidate sequence for $H$ steps to obtain its imagined future trajectory. In the \textit{revision} stage, the value function $H_{\mathrm{UMM}}$ evaluates the terminal imagined states and performs a $K$-way selection to identify the most promising action sequence. Crucially, the selected sequence $a_{0:H-1}$ is not executed in full. Instead, the agent commits only to its first $L$ actions $a_{0:L-1}$, receives a new observation, and replans from the updated state. This receding-horizon closed-loop design preserves the long-horizon lookahead of model-based planning while enabling continual feedback correction, making it substantially better suited to partially observable tasks.

Notably, unlike existing approaches that finetune VLMs for decision-making~\citep{palm-e, enbodiedgpt}, our framework does not rely on expert demonstration datasets. The forward and inverse dynamics models can be finetuned on any available transition data (expert or non-expert). The policy component requires no finetuning, while the heuristic value function for open-loop planning only necessitates labeled data to learn to estimate the number of steps to a goal.

\vspace{-4mm}

\section{Experiments}
\begin{table*}[t]
\centering
\small
\caption{Success rates of planning with different methods.}
\label{table:main_results}
\resizebox{\textwidth}{!}{
\begin{tabular}{lccccccc}
\toprule
\textbf{Model} & \multicolumn{2}{>{\columncolor{cyan!20}}c}{\textbf{2D Navigation}} & \multicolumn{2}{>{\columncolor{green!20}}c}{\textbf{Manipulation}} & \multicolumn{2}{>{\columncolor{purple!20}}c}{\textbf{3D Navigation}} & \textbf{Average} \\
\cmidrule(lr){2-3}\cmidrule(lr){4-5}\cmidrule(lr){6-7}
& \textbf{Frozen Lake}
& \textbf{Mini-BEHAVIOR}
& \textbf{Language Table}
& \textbf{Real World}
& \textbf{Active Recognition}
& \textbf{Image-Goal Navigation}
& \\
\midrule
\multicolumn{8}{l}{\textbf{Closed-Source}} \\
\midrule
GPT-5                 & 0.08 & 0.04 & 0.00 & 0.10 & 0.14 & 0.38 & 0.12 \\
GPT-5-Thinking        & 0.44 & 0.30 & 0.82 & \textbf{0.80} & 0.62 & 0.53 & 0.59 \\
GPT-5-Thinking-Tool   & \textbf{0.98} & 0.68 & \textbf{0.90} & \textbf{0.80} & \textbf{0.66} & \textbf{0.58} & 0.77 \\
\midrule
\multicolumn{8}{l}{\textbf{Open-Source (three training runs)}} \\
\midrule
Qwen2.5-VL-7B-Ins            & 0.33 & 0.00 & 0.14 & 0.07 & 0.23 & 0.07 & 0.14 \\
Qwen2.5-VL-7B-Ins-CoT        & 0.37 & 0.15 & 0.36 & 0.22 & 0.27 & 0.19 & 0.26 \\
Qwen2.5-VL-32B-Ins           & 0.43 & 0.07 & 0.28 & 0.18 & 0.42 & 0.13 & 0.25 \\
Qwen2.5-VL-32B-Ins-CoT       & 0.49 & 0.51 & 0.57 & 0.33 & 0.46 & 0.27 & 0.44 \\
\rowcolor{gray!20}
BAGEL-VLM (baseline, 7B)         & 0.38 & 0.01 & 0.23 & 0.09 & 0.41 & 0.16 & 0.21 \\
\rowcolor{gray!20}
BAGEL-VLM-CoT (baseline, 7B)     & 0.43 & 0.48 & 0.51 & 0.26 & 0.49 & 0.29 & 0.41 \\
VLP                                & 0.94 & 0.76 & 0.28 & 0.43 & 0.57 & 0.39 & 0.56 \\
\midrule
\multicolumn{8}{l}{\textbf{Ours}} \\
\midrule
\rowcolor{orange!20}
Uni-Plan (14B-A7B)           & 0.95 & \textbf{0.83}  & 0.73  & 0.63  & 0.61 & 0.43 & 0.70 \\
\bottomrule
\end{tabular}
}
\end{table*}

In this section, we empirically validate three claims:
\begin{itemize}
    \item Compared to VLM-based planning methods, our method is better at embodied decision-making tasks, including both navigation and manipulation tasks.
    \item The strong decision-making ability is rooted in the fact that the fine-tuned Unified Multimodal Model (UMM) serves as a highly generalizable dynamics model, and is further strengthened by our proposed \textit{self-discriminated filtering}, which rejects implausible transitions to improve prediction accuracy.
    \item Our approach demonstrates superior data scalability in two aspects: it requires no expert demonstrations for finetuning and achieves stronger performance than VLMs when trained with the same amount of data.
\end{itemize}

\subsection{Evaluation of Planning Ability under OOD Environments}
\label{sec:3.1}
\paragraph{Tasks.}
To comprehensively evaluate the planning ability of different models, we consider six tasks spanning three categories. \textbf{2D navigation}: (i) \textit{FrozenLake}, a maze-like environment where the agent must plan a path from a start location to a goal while avoiding traps; and (ii) \textit{Mini-BEHAVIOR}, a series of grid-world embodied AI tasks in which the agent navigates to satisfy goal conditions such as picking up a target object and placing it at a designated location. These tasks primarily evaluate decision-making under discrete spatial constraints. \textbf{Manipulation}: (iii) \textit{Language Table}, a tabletop object-rearrangement task where the agent manipulates blocks to match a target configuration shown in an image; and (iv) \textit{Real World}, a more challenging real-world object-rearrangement setting with unseen objects and containers, requiring accurate recognition and coherent multi-step planning. \textbf{3D navigation}: we further adopt two tasks from World-in-World~\citep{worldinworld}: (v) \textit{Active Recognition}, where the agent must actively move in a 3D scene to collect informative views before identifying the target; and (vi) \textit{Image-Goal Navigation}, where the agent navigates to reach a target position given a single reference image. Both are challenging due to \textit{partial observation}, since the agent only sees a limited egocentric view at each step. For planning, we use the open-loop planner on the 2D navigation and manipulation tasks, and the closed-loop planner on the 3D navigation tasks.

\vspace{-2mm}
\paragraph{Training \& Test Sets.}
We evaluate all tasks under explicit out-of-distribution (OOD) train/test splits, where the test environments differ from training in scene layouts, object configurations, or goal specifications depending on the domain. For all simulated tasks, we use 500 trajectories for finetuning, while the real-world task uses 200 trajectories due to data-collection cost. Notably, the datasets used to finetune the VLM baselines are expert demonstrations, whereas our approach requires no expert data and instead uses an equal amount of non-expert trajectories for a fair comparison. Full training/test-set details, dataset statistics, and collection procedures are provided in Appendix~\ref{appendix:data_collection}.

\begin{figure*}[t]
    \centering
    \vspace{-7mm}
    \includegraphics[width=\textwidth]{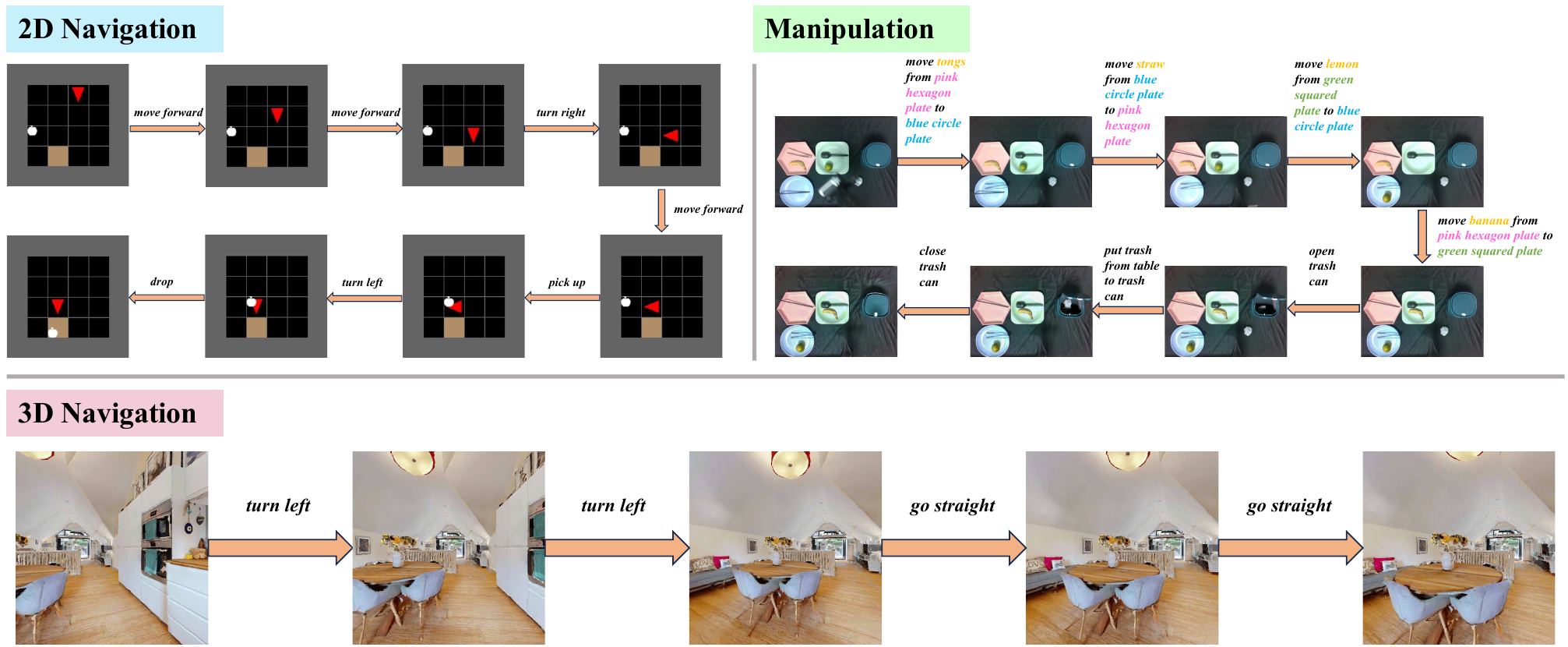}
    \caption{Planning visualizations of Uni-Plan on three task categories: 2D navigation, manipulation, and 3D navigation.}
    \label{fig:planning_vis}
    \label{fig:real_world_task_vis}
    \vspace{-3mm}
\end{figure*}

\vspace{-3mm}
\paragraph{Baselines.} We adopt direct VLM-based planners as our main baselines, following recent findings that strong VLMs can achieve state-of-the-art planning performance without the auxiliary visual grounding modules required by earlier frameworks~\citep{look-before-you-leap}.

For open-source baselines, we include the mainstream VLM Qwen2.5-VL~\citep{qwen2.5vl} and additionally compare BAGEL-VLM, which is BAGEL restricted to its VLM-only mode and thus serving as a natural ablation of our method. For these models, we consider both \textit{non-CoT} and \textit{CoT} variants:
\begin{itemize}
  \item \textbf{Non-CoT version}: The model is fine-tuned using only the final answers so that it outputs a complete plan directly.
  \item \textbf{CoT version}: The model is additionally provided with rationales~\citep{cot} during fine-tuning, enabling it to think step-by-step at inference.
\end{itemize}

For closed-source VLMs\footnote{Because the closed-source models cannot be
finetuned directly, we employ few-shot prompting for in-context learning instead.}, we choose GPT-5~\citep{gpt-5} and its variants:
\begin{itemize}
  \item \textbf{GPT-5}: A standard chat model with relatively limited reasoning ability. We prompt it to think step by step to produce a plan.
  \item \textbf{GPT-5-Thinking}: A stronger model trained to reason through reinforcement learning.
  \item \textbf{GPT-5-Thinking-Tool}: Extends GPT-5-Thinking with a code interpreter, allowing it to write and execute code for better problem solving.
\end{itemize}

Additionally, we include \textit{Video Language Planning (VLP)}~\citep{vlp}, which is the planning baseline most closely related to our method: it uses a VLM as the policy and value function while employing a video generation model as the dynamics model, but does not include our self-discriminated filtering. Since the original VLP implementation is not open-sourced, we reproduce it using Qwen2.5-VL-7B as the VLM and Wan2.2~\citep{wan} as the video generation model.

\vspace{-2mm}
\paragraph{Main Results.}
Table~\ref{table:main_results} reports the planning success rates of all evaluated methods, measured by invoking the corresponding low-level policies to execute the generated plans in the environments. Our approach consistently outperforms open-source VLM baselines by a substantial margin. In particular, compared with the BAGEL-VLM, our method achieves about 50\% higher success rates than its non-CoT variant and nearly 30\% higher than its CoT variant across all reported tasks. This directly highlights the superiority of our planning system over traditional chain-of-thought (CoT) reasoning based solely on text, as both approaches share the same underlying model, yet ours more effectively exploits BAGEL’s capabilities. We provide a detailed analysis of VLM-based planning failure cases in the Appendix~\ref{appendix:vlm_analysis}. When compared with the advanced closed-source model, our method still exhibits comparable performance, which is a promising result given that GPT-5-Thinking-Tool is a significantly larger model and possesses broader knowledge. This also indicates that, beyond the use of external tools to enhance a model’s visual reasoning capability, leveraging the UMM's inherent multimodal generation ability can likewise significantly strengthen reasoning performance, providing a new perspective for improving visual reasoning in the future.

Compared with VLP, our method shows a clear advantage on the two manipulation benchmarks, where precise object-centric reasoning is especially important; we attribute this gain to the stronger dynamics consistency enabled by self-discriminated filtering. In addition, Uni-Plan is substantially more efficient at inference: because it uses a single model and only needs to generate images rather than videos, it achieves roughly a 10$\times$ speedup over VLP, as shown in Appendix Table~\ref{tab:language_table_inference_comparison}.

\vspace{-4mm}
\paragraph{Planning Visualizations.}
We further visualize representative rollouts produced by Uni-Plan on three task categories: 2D navigation, manipulation, and 3D navigation. As shown in Figure~\ref{fig:planning_vis}, the planner generates coherent action sequences whose imagined trajectories remain consistent with the task dynamics across these diverse domains. These examples provide qualitative evidence that Uni-Plan maintains both action plausibility and dynamics consistency across substantially different decision-making settings. We refer readers to Appendix~\ref{appendix:vis_BAGEL} for more visualizations.

\vspace{-2mm}
\subsection{Strong Dynamics Model as the Core of Strong Decision-making Ability}
\label{sec:3.2}
Unlike chain-of-thought (CoT) reasoning, which instinctively generates a reasoning trace in an autoregressive manner, our approach performs beam search over several sampled action candidates, allowing it to escape the limitations of a fixed policy and adapt to novel situations. However, this advantage hinges on the model’s ability to accurately predict the outcomes of different actions.

 \begin{figure*}[ht]
    \vspace{-3mm}
    \centering
    \begin{minipage}{0.48\textwidth}
        \centering
        \includegraphics[width=\textwidth]{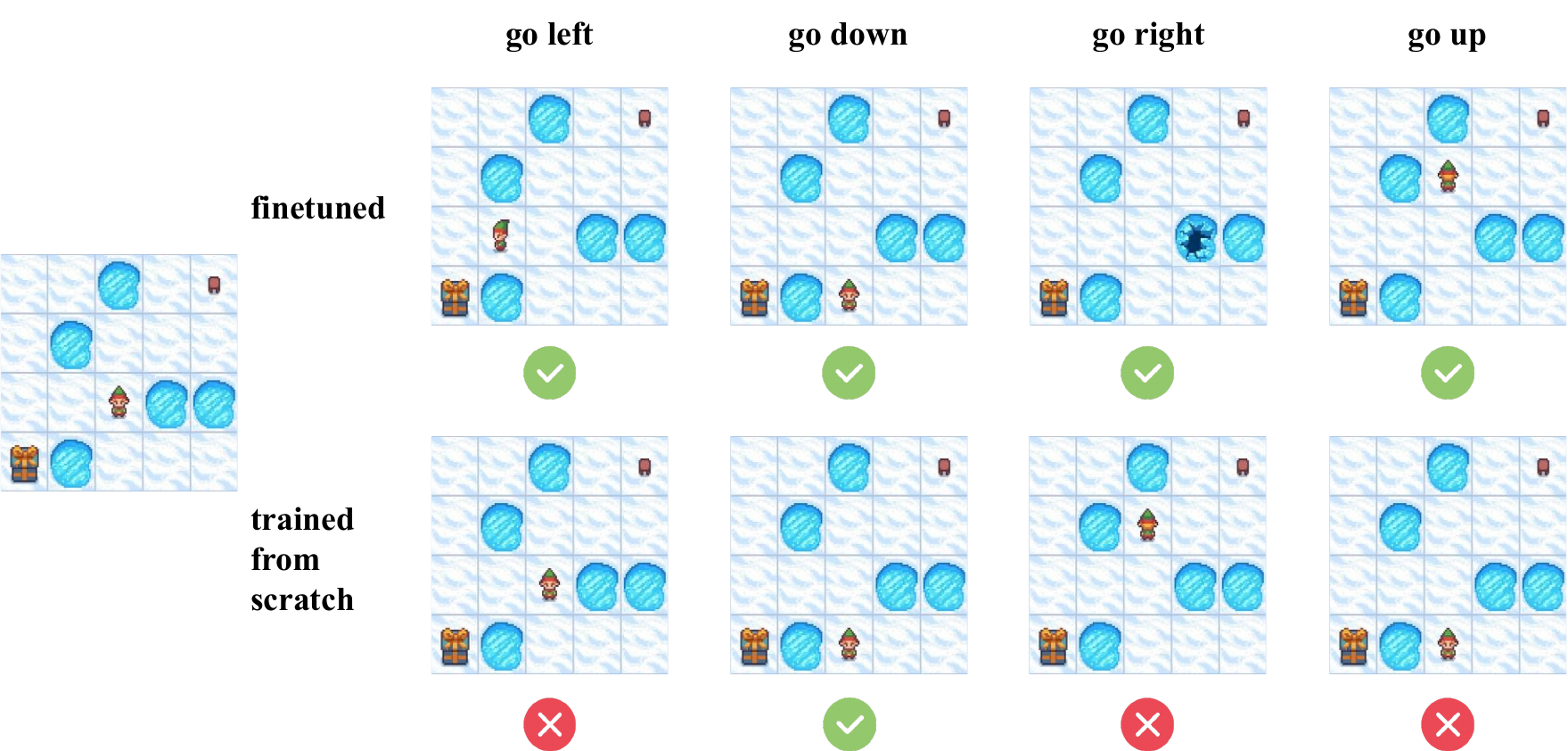}
        \caption{Illustrations of predictions on an OOD case by finetuned BAGEL and BAGEL trained from scratch as dynamics models.}
        \label{fig:ood_predictions}
    \end{minipage}
    \hfill
    \begin{minipage}{0.48\textwidth}
        \centering
        \includegraphics[width=\textwidth]{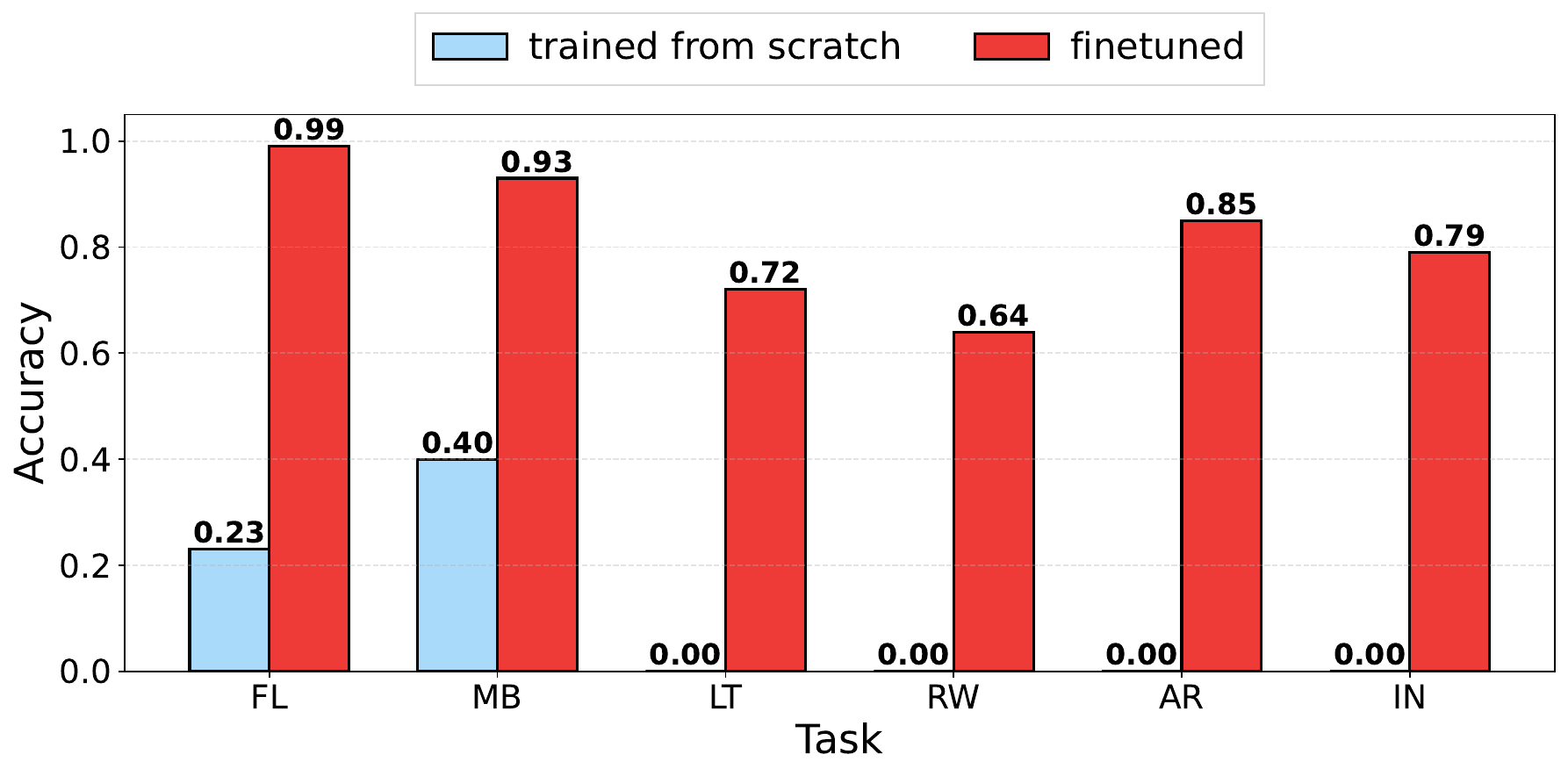}
        \caption{Quantitative comparison between fintuned BAGEL and BAGEL trained from scratch as dynamics models.}
        \label{fig:ood_predictions_quant}
    \end{minipage}
    \vspace{-3mm}
\end{figure*}

We show that the finetuned BAGEL does serve as a strong dynamics model. Figure~\ref{fig:ood_predictions} illustrates examples of transition predictions in an OOD case of the FrozenLake task, where both the maze layout (trap positions, start point, and goal) and the grid size differ from those in the training set. We observe that finetuned BAGEL demonstrates strong generalization ability, producing correct predictions for all possible actions. In contrast, BAGEL trained from scratch performs poorly on this OOD layout. As further supported by the quantitative analysis in Figure~\ref{fig:ood_predictions_quant}, where we evaluate models on 100 OOD scenarios per task by measuring the accuracy of transition predictions, the finetuned model achieves substantially higher prediction accuracy than the model trained from scratch. These results indicate that BAGEL’s strong generalization stems from its pretrained model and that, with few-shot finetuning, it can serve as a reliable dynamics model for downstream tasks. Additional qualitative comparisons between the two models are provided in Appendix~\ref{appendix:vis_BAGEL}.

\begin{figure*}[h]
    \centering
    \begin{minipage}{0.48\textwidth}
        \centering
        \includegraphics[width=\linewidth]{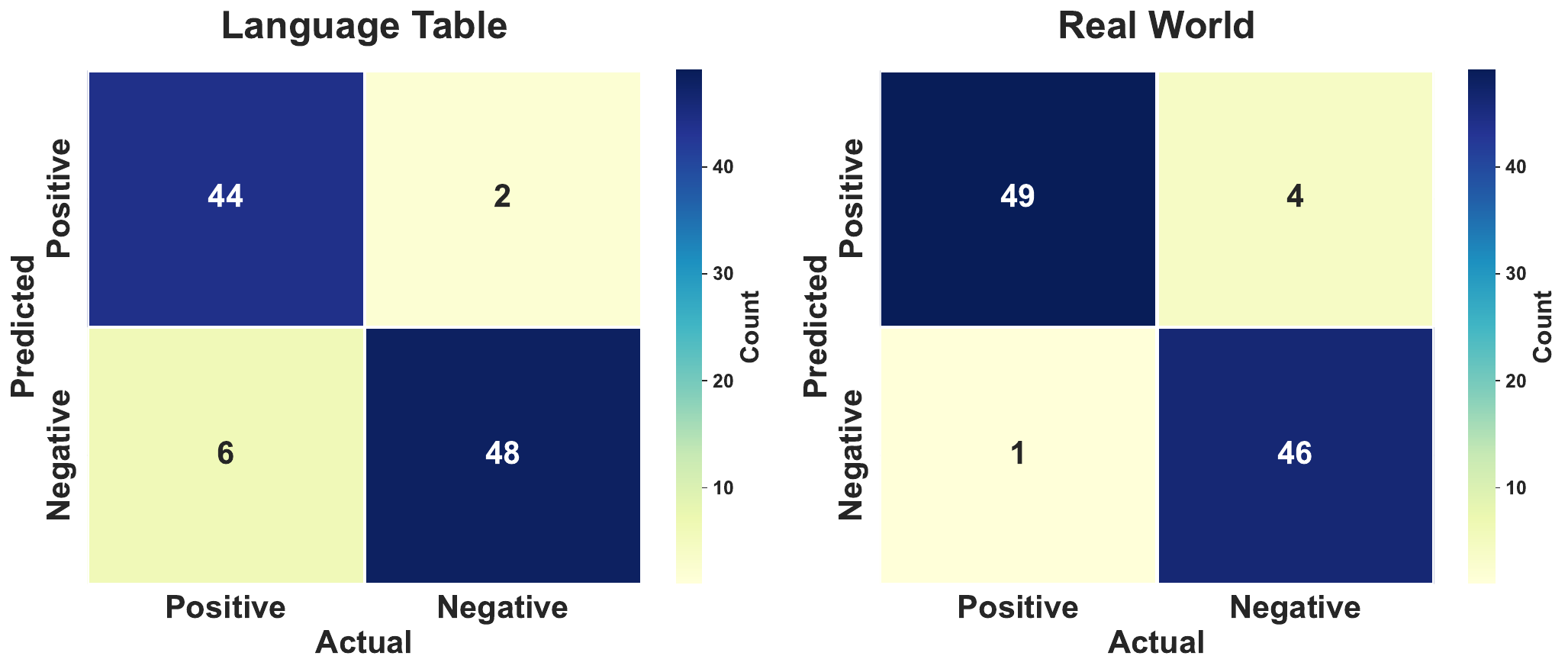}
        \caption{Confusion matrices of predictions by self-discriminated filtering.}
        \label{fig:confusion_matrices}
    \end{minipage}
    \hfill
    \begin{minipage}{0.48\textwidth}
        \captionsetup{type=table}
        \caption{Success rates under different beam-search hyperparameters on Language Table.}
        \label{tab:beam_search_ablation}
        \centering
\small
\setlength{\tabcolsep}{5pt}
\begin{adjustbox}{max width=\linewidth}
\begin{tabular}{ccccc}
    \toprule
    \multicolumn{2}{c}{\textbf{Search Hypers}} & \multicolumn{3}{c}{\textbf{Method}} \\
    \cmidrule(r){1-2} \cmidrule(l){3-5}
    \textbf{Beams} & \textbf{Action Branch} & \textbf{Uni-Plan} & \textbf{w/o filtering} & \textbf{w/o pretraining} \\
    \midrule
    1 & 1 & 0.27 & 0.23 & 0.00 \\
    1 & 4 & 0.55 & 0.20 & 0.00 \\
    2 & 4 & 0.63 & 0.25 & 0.00 \\
    \bottomrule
\end{tabular}
\end{adjustbox}
    \end{minipage}
\end{figure*}

\begin{figure*}[t]
    \vspace{-7mm}
    \centering
    \includegraphics[width=\linewidth]{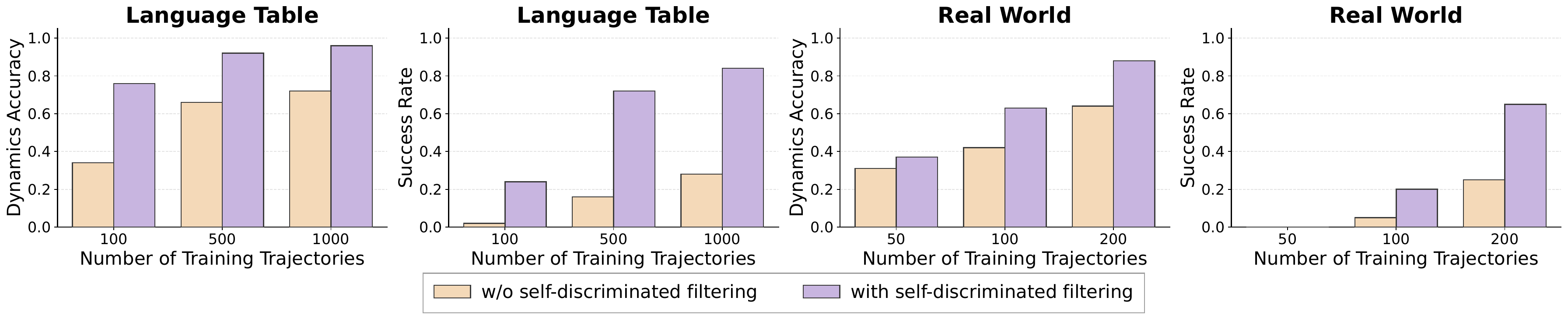}
    \caption{Ablations on self-discriminated filtering.}
    \label{fig:sdf_ablation}
\end{figure*}

The finetuned dynamics model is further enhanced by our proposed \textit{self-discriminated filtering}. Before evaluating its impact, we first verify that this technique is capable of reliably distinguishing correct dynamics predictions from incorrect ones. As shown by confusion matrices in Figure~\ref{fig:confusion_matrices}, it achieves high accuracy and recall with a low false-positive rate, indicating its strong ability to judge the correctness of its own predicted transitions.
Subsequently, we conduct an ablation study to investigate its influence on planning. Figure~\ref{fig:sdf_ablation} reports the accuracy of dynamics predictions and the planning success rates with and without the filtering. As shown in Figure~\ref{fig:sdf_ablation}, the self-discriminated filtering effectively reduces prediction errors in dynamics and, as a result, substantially improves planning success rates.

To further address whether our gains could be explained merely by stronger planning rather than a stronger dynamics model, we vary the beam-search configuration and compare three model variants under the same search budgets. As shown in Table~\ref{tab:beam_search_ablation}, increasing the search width does improve the full Uni-Plan system, but the same increase does not rescue weak dynamics models. In particular, the variant without pretraining remains at zero success rate for all tested beam-search settings, and the variant without filtering shows only limited gains. These results indicate that planning is indeed beneficial, but only when built on top of a sufficiently strong dynamics model; better search alone cannot compensate for poor transition modeling.

\vspace{-2mm}
\subsection{Data Scaling without Expert Demonstrations}
\label{sec:3.3}
\begin{figure*}[ht]
    \centering
    \includegraphics[width=1.0\linewidth]{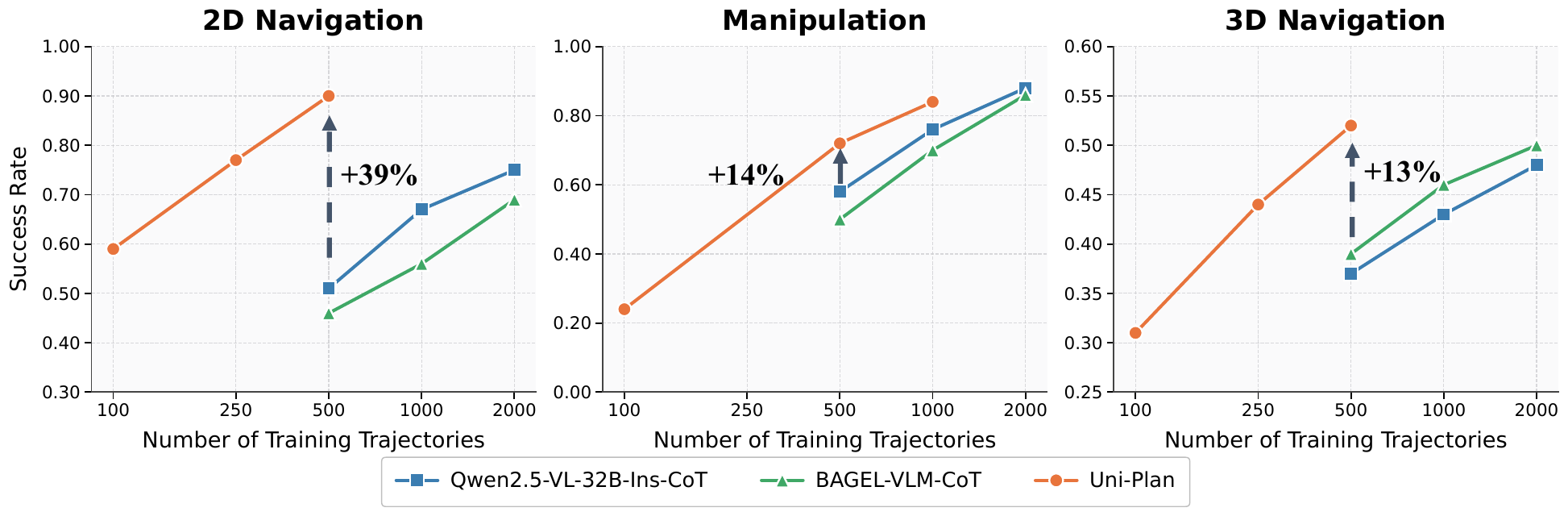}
    \vspace{-3mm}
    \caption{Data scaling trends of two VLM methods: Qwen2.5-VL-32B-Ins-CoT and BAGEL-VLM-CoT, and our method Uni-Plan.}
    \label{fig:data_scaling}
    \vspace{-3mm}
\end{figure*}

In this section, we examine the data scalability of our planning system.
While existing VLM baselines require expert-collected datasets to adapt their policies \citep{palm-e,enbodiedgpt}, our framework can be trained effectively on non-expert trajectories, where actions may be suboptimal. Figure~\ref{fig:data_scaling} compares scaling trends for Qwen2.5-VL-32B-Ins-CoT, BAGEL-VLM-CoT, and Uni-Plan. Despite relying only on non-expert data, Uni-Plan consistently achieves higher performance with the same amount of data. In particular, just 500 trajectories are sufficient for Uni-Plan to reach strong performance, whereas VLMs fail to achieve competitive results even with four times as much data on 2D navigation tasks.

\vspace{-2mm}
\section{Related Work}

Our work intersects with several research areas, including decision-making with language models, thinking with images, and self-verification. We provide a comprehensive discussion below.

\vspace{-3mm}
\paragraph{Decision-Making with Language Models.} 
Motivated by the strong reasoning capabilities of large language models (LLMs) and vision–language models (VLMs), many studies have explored their application to decision-making. \cite{llm-zero-shot-planner} demonstrate that LLMs can serve as zero-shot planners, decomposing high-level tasks into mid-level plans via prompting. \cite{saycan} augment LLMs with affordances to ground them in real-world robotic tasks. \cite{inner-monologue} further incorporate environment feedback to form an inner monologue, enabling richer planning and control. However, all of these approaches operate solely in the text modality and thus lack grounded perception. To overcome this limitation, several works leverage VLMs for decision-making. For instance, \cite{palm-e} and \cite{enbodiedgpt} combine LLMs with vision encoders to form VLMs that, after finetuning on embodied datasets, can handle a wide range of embodied planning tasks. \cite{look-before-you-leap} show that advanced closed-source VLMs such as GPT-4V can solve many open-world manipulation tasks without finetuning. Beyond manipulation, \cite{navid} demonstrate the use of VLMs for vision-and-language navigation. While these methods have unique advantages, they generally reduce decision-making to treating an LLM/VLM as a policy, and therefore lack counterfactual reasoning ability. In contrast, another line of work formulates decision-making as planning with world models. \cite{rap} and \cite{gpt-mcts} repurpose LLMs as both world models and reasoning agents, incorporating principled planning algorithms such as Monte Carlo Tree Search for strategic exploration. More recently, \cite{vlp} introduce video language planning (VLP), a framework for complex visual tasks in which VLMs act as policies and value functions, and text-to-video models serve as dynamics models. VLP is most relevant to our work since both VLP and our method employ beam search for visual task planning. Nevertheless, our approach differs in key aspects: (i) we unify all roles within a single model, whereas VLP requires two separate models, making our method more efficient at inference; (ii) our model can act as a self-discriminator to reduce hallucinations; and (iii) we demonstrate superior data scalability compared with VLM-based planning.

\vspace{-3mm}
\paragraph{Thinking with Images.} 
The driving idea of our work is that incorporating images into the thinking process can enhance reasoning ability. Many related studies share the same insight and demonstrate effectiveness on reasoning tasks such as mathematics or VQA. \cite{visual-sketchpad} propose the visual chain-of-thought, which generates Python code to invoke external tools for sketching. \cite{image-of-thought} similarly leverage tools for image manipulation to create visual rationales in chain-of-thought reasoning. These methods, however, rely on external modules for image generation. A more promising direction is native multimodal reasoning. For example, \cite{mvot} finetune a unified multimodal model for multimodal visualization-of-thought, enabling the UMM to produce visualizations of their reasoning traces. \cite{thinking-with-images} propose iterative refinement of image generation through visual reasoning. Despite these advances, existing approaches use UMMs mainly for visualizing reasoning traces or refining generations, rather than for more sophisticated decision-making as in our work.

\vspace{-2mm}
\paragraph{Self-Verification.}
Recent studies explore enabling large language models (LLMs) to verify their own outputs. \cite{llm_self_veri} propose a self-verification strategy that allows large language models (LLMs) to reevaluate their own reasoning to improve answer reliability. \cite{self-check} introduce a multi-stage approach that breaks the problem down into a series of simpler tasks and perform step-by-step check. \cite{s2r} train models via reinforcement learning to strengthen both self-verification and self-correction abilities. However, these approaches focus purely on textual reasoning. In contrast, our proposed \textit{self-discriminated filtering} extends self-verification to multimodal dynamics prediction, where a UMM generates candidate next observations and verifies them via inverse-dynamics inference, filtering invalid transitions.

\section{Conclusion and Future Directions}

In this paper, we presented \textit{Uni-Plan}, a planning framework built on Unified Multimodal Models (UMMs) where a single model simultaneously serves as policy, dynamics model, and value function. The central challenge we identified is learning a faithful dynamics model. To address this, we introduced a self-discriminated filtering mechanism that allows the generative model to act as its own discriminator, filtering out invalid dynamics predictions. Experimental results show that Uni-Plan outperforms VLM-based paradigms on embodied decision-making tasks, owing to its capacity to function as a highly generalizable dynamics model further reinforced by our proposed filtering method. Uni-Plan also exhibits strong data scalability, requiring no expert demonstrations for fine-tuning and outperforming VLMs when trained with the same amount of data.  

Unlike prior approaches that use generated images primarily for visualization of reasoning traces, Uni-Plan employs image generation for \textit{counterfactual reasoning}. By simulating multiple possible futures under different actions, the model does not merely illustrate its thought process but actively evaluates alternative trajectories. This constitutes a shift from visual explanation to visual reasoning as computation, where generated images are intermediates in search and decision-making rather than expository artifacts. We view this as a conceptual leap: images here are not outputs to be consumed by humans but internal representations used by the model to reason about the world.  

Our approach also opens up a promising direction for future work. In particular, the current limitation of our framework is that it still relies on task-specific finetuning for each downstream domain. A natural next step is to move toward a zero-shot planner by pretraining the model on substantially broader and more diverse dynamics data, with the goal of learning a more general world model that can transfer across tasks without additional adaptation. We view this direction as an exciting opportunity to further extend the scope of counterfactual visual reasoning for planning.

\normalem
\bibliographystyle{plainnat}
\bibliography{reference}

\newpage
\newpage
\appendix
\section{Brief Introduction to BAGEL}
\label{appendix:intro_to_bagel}
BAGEL~\citep{bagel} is an open-source foundational model that natively supports both multimodal understanding and generation. In this section, we describe its architecture, pretraining data, and the capabilities of the pretrained model.

The backbone of BAGEL is derived from the Qwen2.5 LLM~\citep{qwen2.5}. For visual understanding, it employs a Vision Transformer (ViT) encoder to convert raw pixels into visual tokens. For visual generation, BAGEL first applies a pretrained VAE to map images from pixel space to a latent space, and then adopts Rectified Flow~\citep{flow-matching, rectified-flow} in that latent space to generate images. Text generation is performed autoregressively, whereas image generation proceeds in parallel. In addition, BAGEL adopts a Mixture-of-Transformers (MoT) architecture that uses separate QKV projectors and feed-forward networks (FFNs) for understanding and generation while sharing the same attention layers. Each component is initialized from Qwen2.5-7B, resulting in a total of roughly 14B parameters (only 7B parameters active during inference).

BAGEL is pretrained on interleaved multimodal datasets encompassing multimodal conversation, text-to-image generation, and image manipulation, which enables seamless integration of diverse generative tasks. In the early stages of pretraining, it is primarily trained on simple text-to-image (T2I) and image-to-text (I2T) pairs; later stages introduce high-resolution T2I and I2T pairs as well as interleaved multimodal understanding and generation data. Cross-entropy loss is applied to text tokens, while mean-squared error loss is used for image token generation.

Thanks to this comprehensive training corpus, BAGEL exhibits superior visual understanding and image generation capabilities compared with other leading open-source models~\citep{qwen2.5vl, internvl2.5, seed-x, janus-pro}. More importantly, BAGEL is capable of high-fidelity image editing, which is more challenging than T2I generation because it requires precise control over image details according to textual instructions while maintaining overall visual consistency. This image-editing ability underpins its role as a reliable dynamics model in our planning framework.

\section{Implementation Details}
\subsection{Data Collection}
\label{appendix:data_collection}
\begin{figure}[!t]
    \vspace{-5mm}
    \centering
    \includegraphics[width=\linewidth]{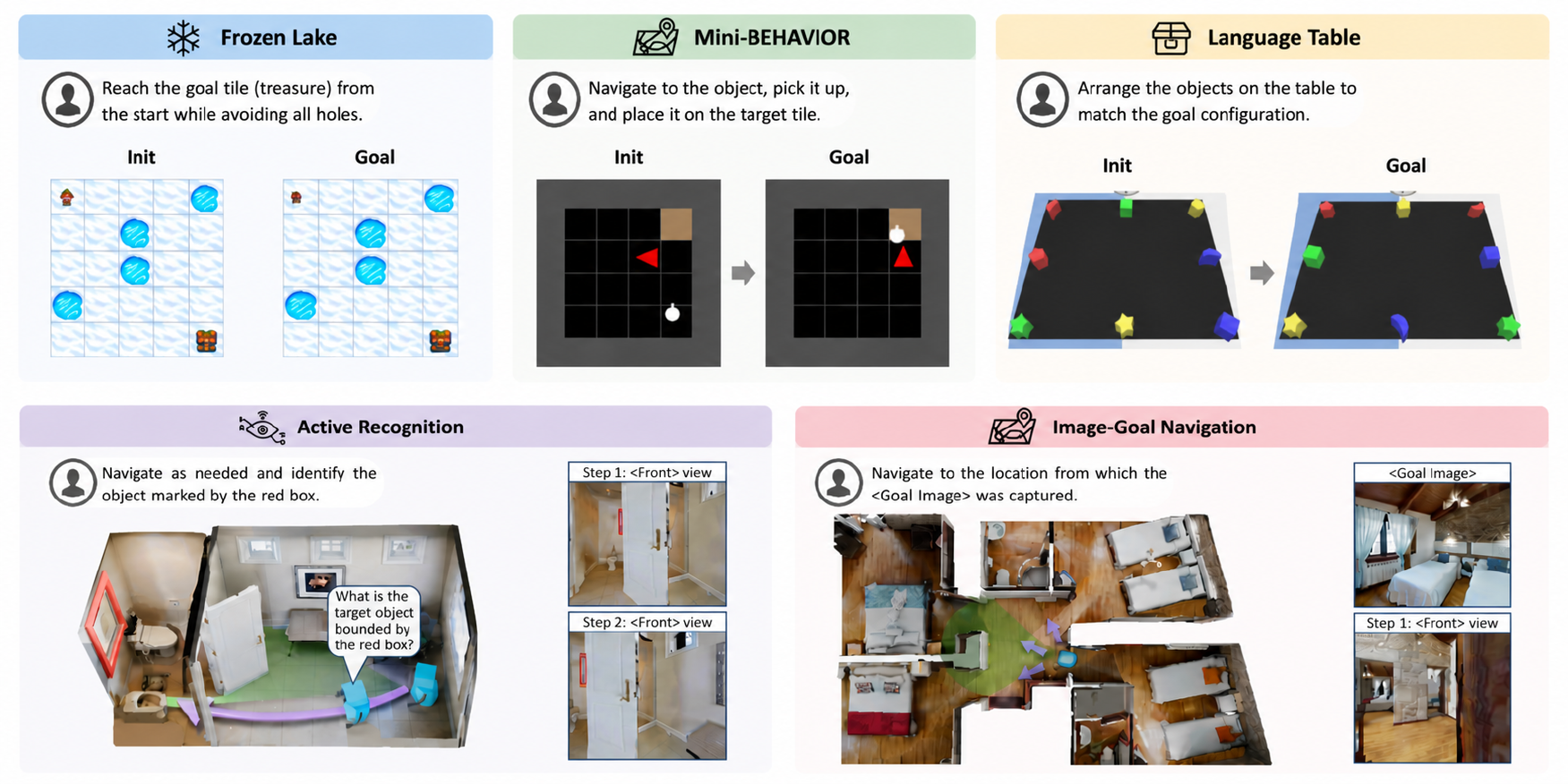}
    \caption{Illustrations of tasks for our evaluation.}
    \label{fig:sim_task_demos}
\end{figure}

\paragraph{Simulation tasks.}
There are five simulated environments in our experiments: \textit{FrozenLake}, \textit{Mini-BEHAVIOR}, \textit{Language Table}, \textit{Active Recognition}, and \textit{Image-Goal Navigation}. We show some illustrations of these tasks in Figure~\ref{fig:sim_task_demos}. For the simulated tasks, we train agents to collect expert trajectories and also randomly sample some non-expert trajectories.

For the train/test split, we explicitly evaluate all simulated tasks under out-of-distribution settings. In \textit{FrozenLake}, the test set contains unseen layouts that differ from training in both map size and trap distribution. In \textit{Mini-BEHAVIOR}, the test set comprises novel maps with different start, object, and goal positions. In \textit{Language Table}, the test set uses distinct block configurations for both the initial and goal states. For the two 3D navigation tasks adopted from World-in-World~\citep{worldinworld}, we let training and test episodes come from disjoint 3D scenes and different goals. We use 500 trajectories for each simulated task.

\begin{figure}[!t]
    \vspace{-5mm}
    \centering
    \includegraphics[width=\linewidth]{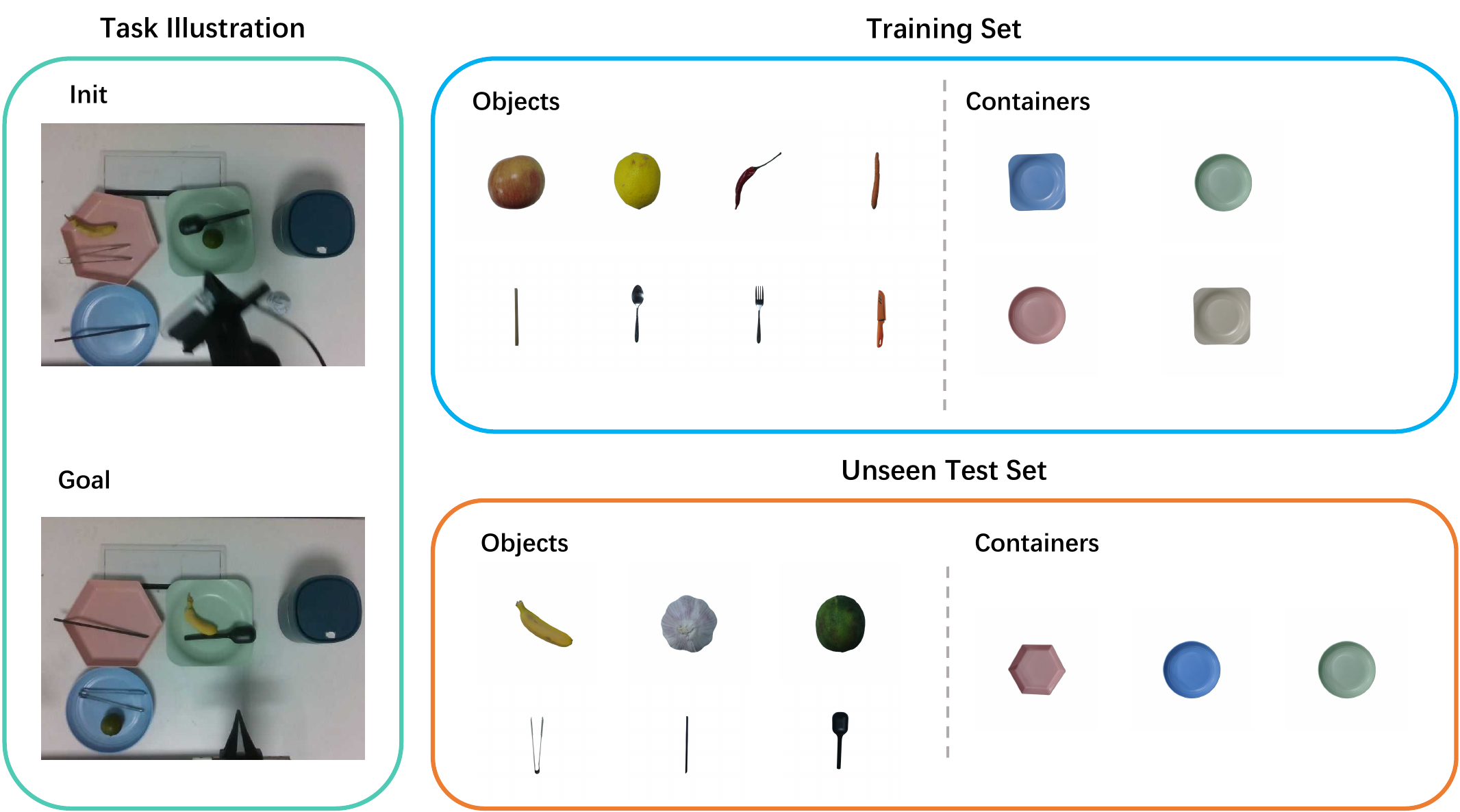}
    \caption{Setting for real-world tasks. The training set contains 8 objects (4 foods and 4 cutlery) and 4 containers, and the test set contains 6 unseen objects and 3 unseen containers.}
    \label{fig:real_world_task_setting}
\end{figure}

\paragraph{Real-world tasks.}
For the real-world task (visualized in Figure~\ref{fig:real_world_task_vis}), we collect 200 full-horizon expert demonstrations via human teleoperation. The task requires the robot to execute a long-horizon sequence of three subtasks to rearrange the objects on the table to the goal:  
\begin{itemize}
\item \textit{Open} or \textit{close the trash can} (if trash-related actions occur).
\item \textit{move X on Y to Z} with varying object-container pairs, where \textit{X} denotes the target object and \textit{Y} and \textit{Z} represent the source and the target containers.
\end{itemize}
As illustrated in Figure~\ref{fig:real_world_task_setting}, we involve unseen objects and containers in the test set for challenging visual discrimination and dynamics prediction. Moreover, when trash manipulation is involved, the robot must open the trash can before the first such action and close it after the last, enforcing stateful environmental awareness.
Overall, this real-world task demands joint competence in fine-grained object and container recognition as well as accurate dynamics prediction.

\paragraph{Overall data statistics.}
We train Uni-Plan on mixed expert and randomly sampled data while training Qwen2.5-VL~\citep{qwen2.5vl} on pure expert data, and keep the same amount of data during training. For all simulated tasks, we use 500 trajectories; for the real-world task, we use 200 trajectories due to the higher cost of teleoperated data collection. Dataset statistics are presented in Table~\ref{table:data_statistics}.

\noindent
\begin{minipage}[t]{0.48\linewidth}
  \centering
  \captionof{table}{Dataset statistics for different tasks.}
  \label{table:data_statistics}
  \resizebox{\linewidth}{!}{
    \begin{tabular}{lccc}
      \toprule
      Tasks & Num Train Trajs & Num Test Trajs & Avg Length \\
      \midrule
      FrozenLake & 500 & 50 & 7.3 \\
      Mini-BEHAVIOR & 500 & 50 & 7.7 \\
      Language Table & 500 & 50 & 8.8 \\
      Real-world Task & 200 & 20 & 8.2 \\
      Active Recognition & 500 & 50 & 9.2 \\
      Image-Goal Navigation & 500 & 50 & 58.4 \\
      \bottomrule
    \end{tabular}
  }
\end{minipage}
\hfill
\begin{minipage}[t]{0.48\linewidth}
  \centering
  \captionof{table}{Hyperparameters of planning.}
  \label{tab:beam_search_hypers}
  \resizebox{\linewidth}{!}{
    \begin{tabular}{lccc}
      \toprule
      \multicolumn{4}{c}{\textbf{Open-loop planning}} \\
      \midrule
      Task & Beam Size & Action Branch & Dynamics Branch \\
      \midrule
      FrozenLake     & 2 & 4 & 1 \\
      Mini-BEHAVIOR  & 2 & 5 & 1 \\
      Language Table & 2 & 4 & 4 \\
      Real World     & 2 & 4 & 8 \\
      \midrule
      \multicolumn{4}{c}{\textbf{Closed-loop planning}} \\
      \midrule
      Task & Planning Horizon & Execution Horizon & Num Proposals \\
      \midrule
      Active Recognition & 4 & 4 & 2 \\
      Image-Goal Navigation & 5 & 3 & 3 \\
      \bottomrule
    \end{tabular}
  }
\end{minipage}

\subsection{Implementation Details of Uni-Plan}
\label{appendix:impl_details}
For completeness, we provide pseudocode for the two planning variants used in Uni-Plan. Algorithm~\ref{alg:open_loop_appendix} summarizes the open-loop planner used in fully observable environments, where beam search is performed over imagined future states. Algorithm~\ref{alg:closed_loop_appendix} summarizes the closed-loop planner used in partially observable environments, where the model repeatedly proposes candidate action sequences, simulates their outcomes, and replans after executing a short action prefix.

\begin{algorithm}[h]
\caption{Uni-Plan (Open-loop)}
\label{alg:open_loop_appendix}
\small
\begin{algorithmic}[1]
\REQUIRE Initial observation $o_0$, goal $g$, action branching factor $K$, dynamics branching factor $D$, beam size $B$, planning horizon $H$
\STATE Initialize beam set $\mathcal{B}_0 \gets \{(o_0, \emptyset)\}$
\FOR{$h = 0$ to $H-1$}
    \STATE Initialize candidate set $\mathcal{C} \gets \emptyset$
    \FOR{each beam $(o_h, a_{0:h-1}) \in \mathcal{B}_h$}
        \STATE Sample $K$ candidate actions $\{a_h^{(k)}\}_{k=1}^{K} \sim \pi_{\mathrm{UMM}}(o_h, g)$
        \FOR{each candidate action $a_h^{(k)}$}
            \STATE Sample $D$ candidate next observations $\{\tilde{o}_{h+1}^{(k,d)}\}_{d=1}^{D} \sim P_{\mathrm{UMM}}(o_h, a_h^{(k)})$
            \STATE Filter $\{\tilde{o}_{h+1}^{(k,d)}\}_{d=1}^{D}$ using self-discriminated filtering to obtain a valid $\hat{o}_{h+1}^{(k)}$
            \STATE Score successor state with $u^{(k)} \gets H_{\mathrm{UMM}}(\hat{o}_{h+1}^{(k)}, g)$
            \STATE Add $(\hat{o}_{h+1}^{(k)}, a_{0:h-1} \cup a_h^{(k)}, u^{(k)})$ to $\mathcal{C}$
        \ENDFOR
    \ENDFOR
    \STATE Keep the top-$B$ candidates in $\mathcal{C}$ as the new beam set $\mathcal{B}_{h+1}$
\ENDFOR
\STATE Return the highest-scoring action sequence in $\mathcal{B}_{H}$
\end{algorithmic}
\end{algorithm}

\begin{algorithm}[h]
\caption{Uni-Plan (Closed-loop)}
\label{alg:closed_loop_appendix}
\small
\begin{algorithmic}[1]
\REQUIRE Current observation $o_0$, goal $g$, number of proposals $K$, planning horizon $H$, execution horizon $L$
\WHILE{goal is not reached}
    \STATE Sample $K$ candidate action sequences $\{a_{0:H-1}^{(k)}\}_{k=1}^{K} \sim \pi_{\mathrm{UMM}}(o_0, g)$
    \FOR{each candidate sequence $a_{0:H-1}^{(k)}$}
        \STATE Set imagined state $\hat{o}_0^{(k)} \gets o_0$
        \FOR{$h = 0$ to $H-1$}
            \STATE Predict next observation $\hat{o}_{h+1}^{(k)} \sim P_{\mathrm{UMM}}(\hat{o}_h^{(k)}, a_h^{(k)})$
        \ENDFOR
    \ENDFOR
    \STATE Select the best terminal state $\hat{o}_H^{\star}$ from $\{\hat{o}_H^{(k)}\}_{k=1}^{K}$ using $H_{\mathrm{UMM}}(\cdot, g)$
    \STATE Let $a_{0:H-1}^{\star}$ denote the action sequence associated with $\hat{o}_H^{\star}$
    \STATE Execute the first $L$ actions of $a_{0:H-1}^{\star}$
    \STATE Receive updated observation $o_0$
\ENDWHILE
\end{algorithmic}
\end{algorithm}

Uni-Plan finetunes BAGEL on each task using 8$\times$H100 GPUs for 3,000 gradient steps with a constant learning rate of $1\mathrm{e}{-6}$, requiring roughly 6 hours of training. During finetuning, the sampling ratio between image-generation data (for the dynamics model) and visual-understanding data (for the policy, value function, and inverse dynamics) is set to 1:1.

As for inference, we list the hyperparameters of Uni-Plan in Table~\ref{tab:beam_search_hypers}. We also present the detailed inference cost in Table~\ref{tab:inference_cost} and compare its end-to-end inference time on \textit{Language Table} with representative VLM baselines in Table~\ref{tab:language_table_inference_comparison}. The comparison reveals a clear test-time scaling trend: methods that spend more inference-time compute generally achieve higher success rates, as seen when moving from standard Qwen2.5-VL to its CoT variant, and from GPT-5 to GPT-5-Thinking-Tool. Uni-Plan follows the same principle, using additional test-time computation for model-based planning and reaching a 0.73 success rate with a 14B-parameter model. 

Compared with the video-based planning method~\citep{vlp}, our method achieves substantially lower inference time, yielding roughly a 10$\times$ speedup. Moreover, faster unified multimodal backbones and more efficient image-generation systems, such as Hyper-BAGEL~\citep{hyper-bagel}, could further reduce this cost substantially; with the reported 22$\times$ generation speedup, the inference latency would be reduced even further.

\noindent
\begin{minipage}[t]{0.48\linewidth}
  \centering
  \captionof{table}{Inference cost of Uni-Plan on one H100 GPU.}
  \label{tab:inference_cost}
  \resizebox{\linewidth}{!}{
    \begin{tabular}{lcc}
      \toprule
      Task & Images/Step & Time/Step \\
      \midrule
      FrozenLake     & 8  & 6s \\
      Mini-BEHAVIOR  & 10 & 8s \\
      Language Table & 32 & 15s \\
      Real World     & 64 & 22s \\
      Active Recognition & 8 & 9s \\
      Image-Goal Navigation & 15 & 16s \\
      \bottomrule
    \end{tabular}
  }
\end{minipage}
\hfill
\begin{minipage}[t]{0.48\linewidth}
  \centering
  \captionof{table}{Inference time and success rate on \textit{Language Table}.}
  \label{tab:language_table_inference_comparison}
  \resizebox{\linewidth}{!}{
    \begin{tabular}{lcc}
      \toprule
      Method & Inference Time & Success Rate \\
      \midrule
      Qwen2.5-VL-7B-Ins & 3s & 0.14 \\
      Qwen2.5-VL-7B-Ins-CoT & 73s & 0.36 \\
      GPT-5 & 7s & 0.00 \\
      GPT-5-Thinking-Tool & 88s & 0.90 \\
      VLP & 1270s & 0.28 \\
      Uni-Plan & 137s & 0.73 \\
      \bottomrule
    \end{tabular}
  }
\end{minipage}
\vspace{5mm}







\newpage
\section{Qualitative Analysis}
\subsection{Qualitative Analysis of Uni-Plan}
\label{appendix:vis_BAGEL}

\begin{figure}[h]
    \centering
    \includegraphics[width=\linewidth]{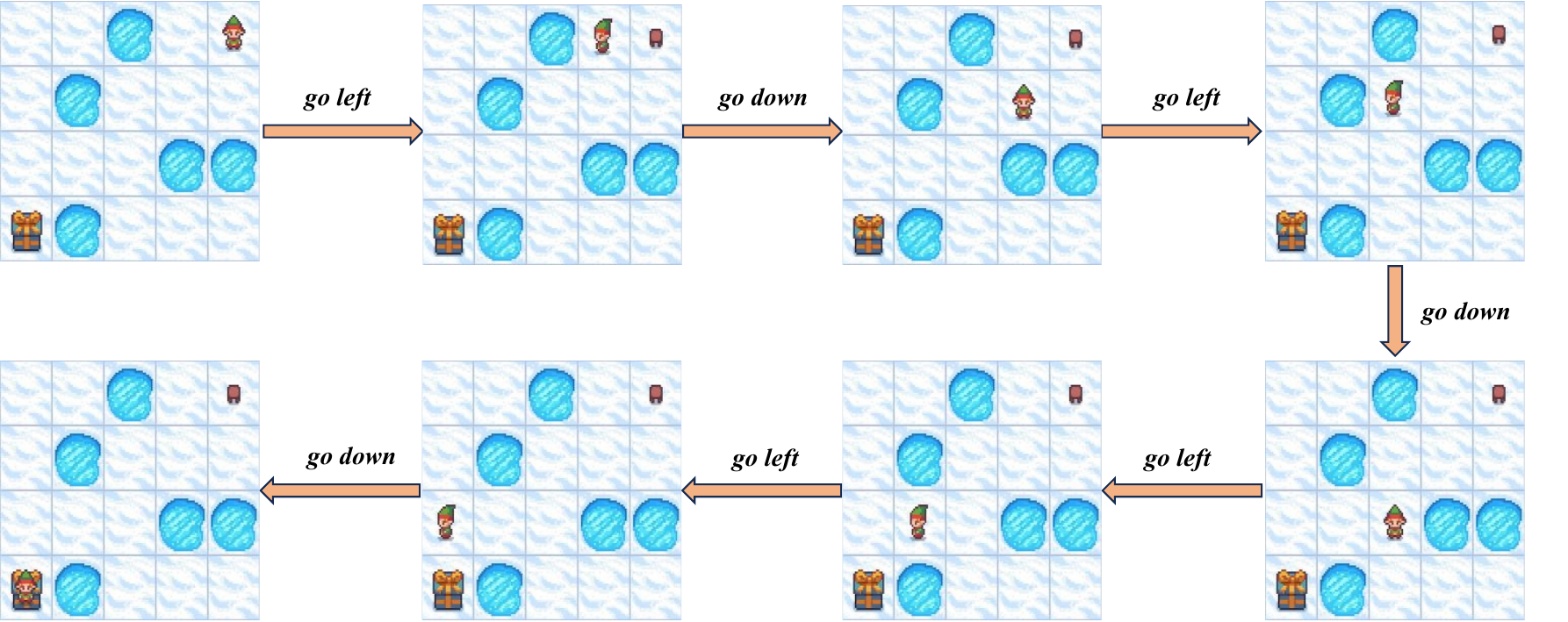}

    \vspace{1mm}\rule{\linewidth}{0.4pt}\vspace{2mm}

    \includegraphics[width=\linewidth]{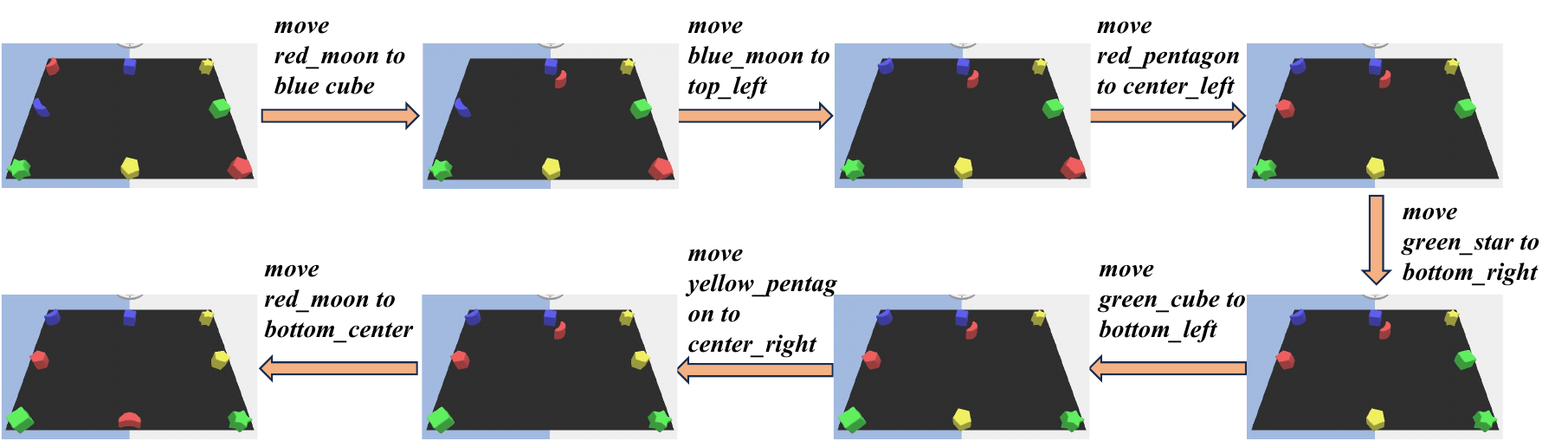}

    \vspace{1mm}\rule{\linewidth}{0.4pt}\vspace{2mm}

    \includegraphics[width=\linewidth]{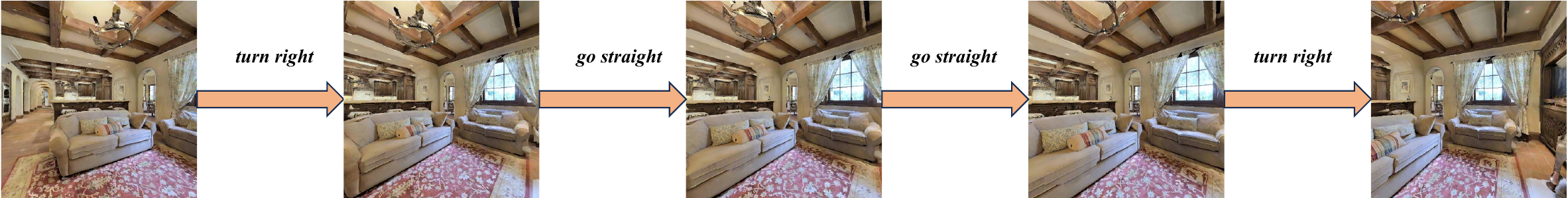}

    \caption{Three planning visualizations with Uni-Plan on FrozenLake, Language Table and Active Recognition.}
    \label{fig:bagel_planning}
\end{figure}

We first present representative successful planning cases of Uni-Plan for each task in Figure~\ref{fig:bagel_planning}.

\newpage

\begin{figure}[!h]
    \vspace{0mm}
    \centering
    \includegraphics[width=\linewidth]{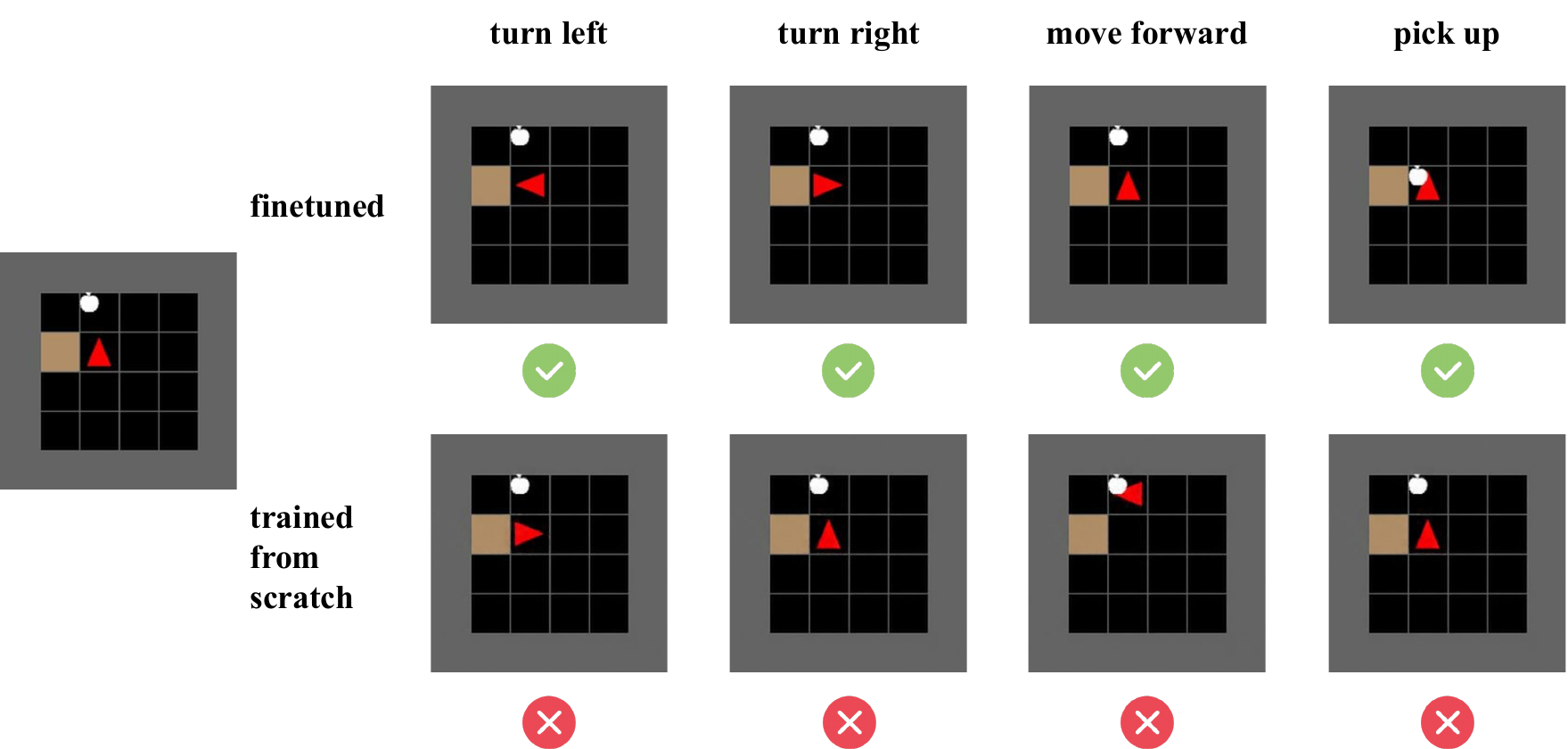}

    \vspace{1mm}\rule{\linewidth}{0.4pt}\vspace{2mm}

    \includegraphics[width=\linewidth]{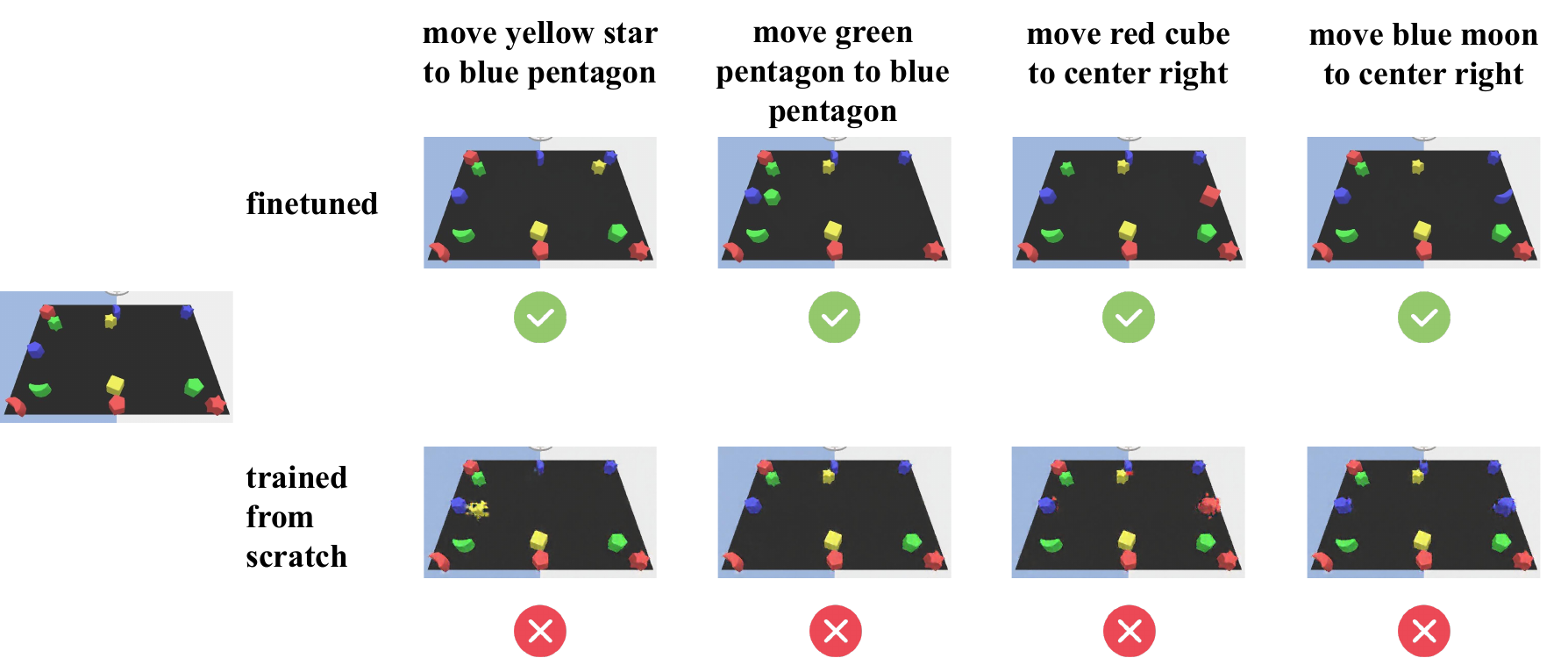}

    \vspace{1mm}\rule{\linewidth}{0.4pt}\vspace{2mm}

    \includegraphics[width=\linewidth]{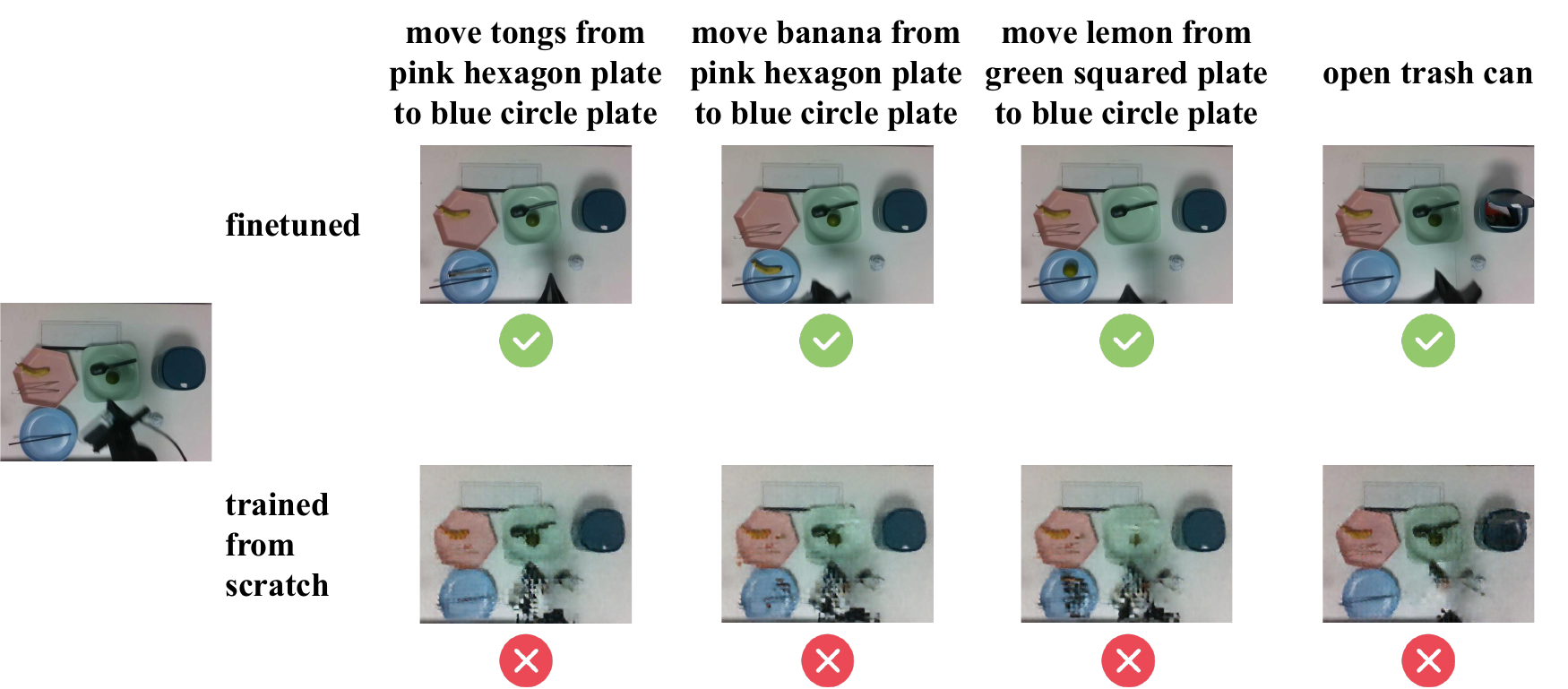}
    \caption{Illustrations of dynamics predictions on unseen samples by finetuned BAGEL and BAGEL trained from scratch}
    \label{fig:additional_dynamics_ood_preds}
\end{figure}

Then, we show additional illustrations of dynamics predictions on unseen samples by finetuned BAGEL and BAGEL trained from scratch in Figure~\ref{fig:additional_dynamics_ood_preds}.

Next, we examine several failure cases, which can be broadly grouped into two categories: \emph{dynamics model errors} and \emph{value function errors}.

\begin{figure}[!h]
    \vspace{-3mm}
    \centering
    \includegraphics[width=\linewidth]{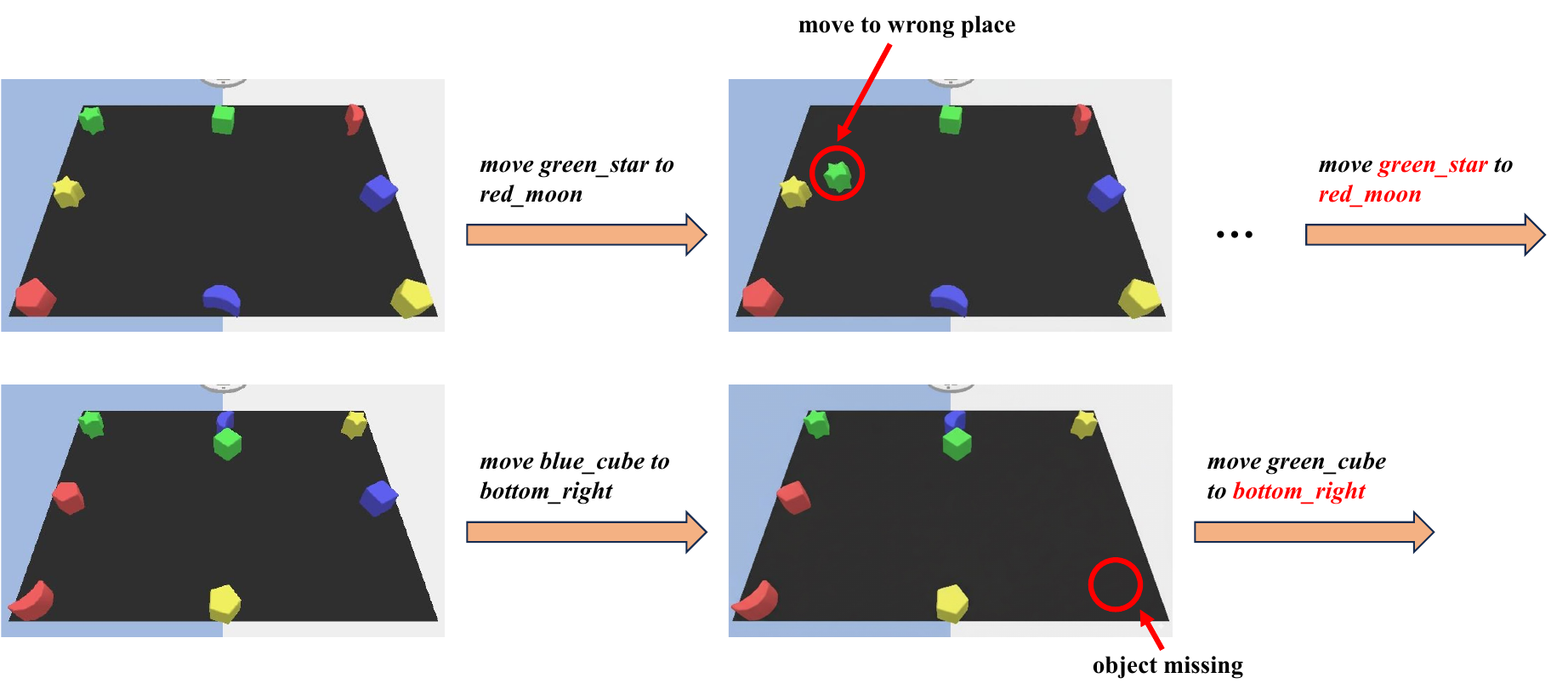}
    \vspace{-3mm}
    \caption{Illustrations of dynamics model errors.}
    \label{fig:dynamics_errors}
\end{figure}

\paragraph{Dynamics Model Errors.}  
Figure~\ref{fig:dynamics_errors} shows failures caused by incorrect dynamics predictions. The top panel illustrates a wrong placement in the transition, which leads to the next action still trying to move other blocks to \textit{red moon}. The bottom panel shows a case where an object is missing from the predicted observation, causing the policy to continue moving other objects toward that location. Although our proposed \textit{self-discriminated filtering} alleviates such issues, these errors can still occur because only a limited number of predictions are sampled for each state–action pair, and we cannot guarantee that at least one valid prediction will be included, especially in low-data regimes.

\begin{figure}[!h]
    \centering
    \includegraphics[width=\linewidth]{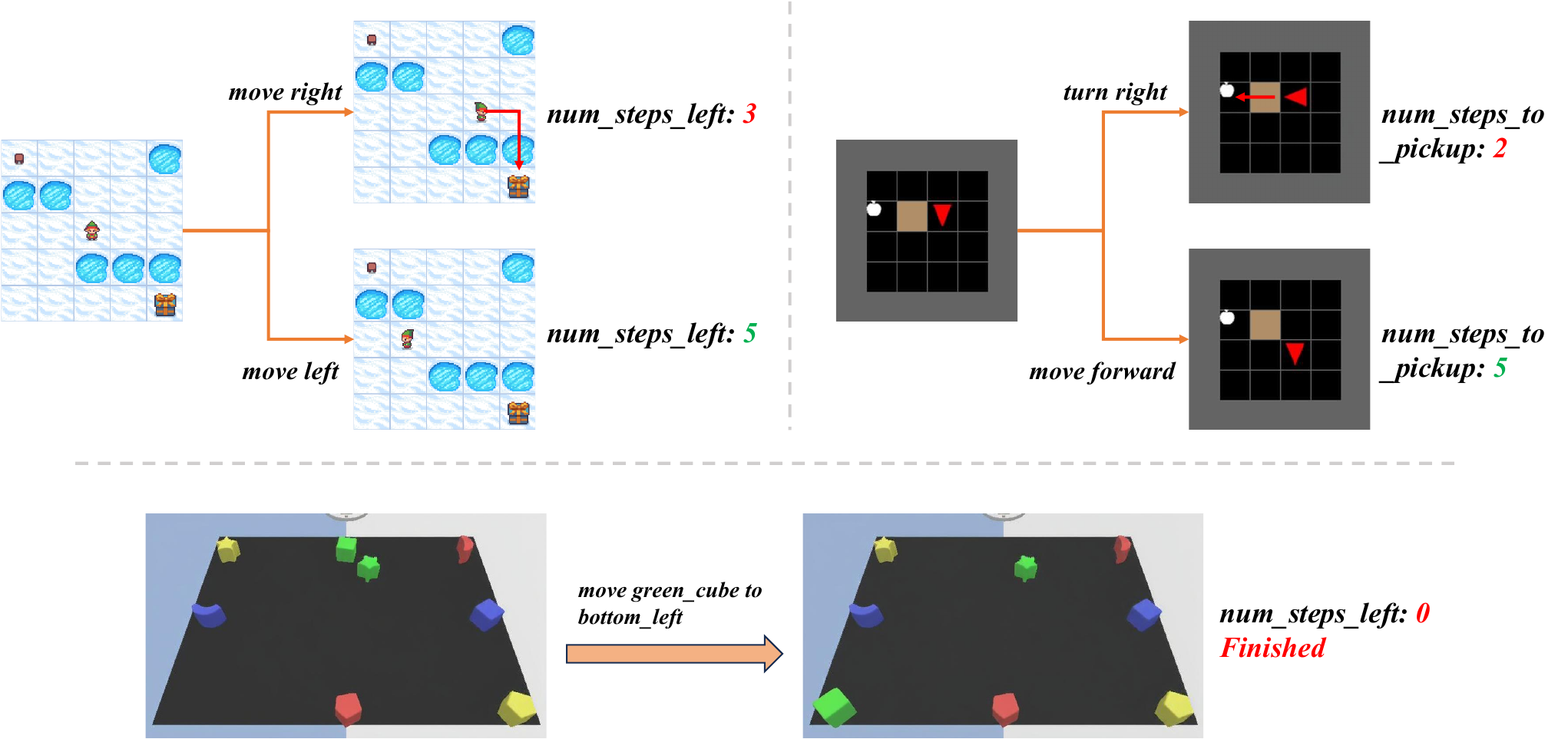}
    \caption{Illustrations of value function errors.}
    \label{fig:value_errors}
\end{figure}

\paragraph{Value Function Errors.}
Figure~\ref{fig:value_errors} depicts failures arising from inaccurate value estimates, which frequently occur in data-scarce regimes, such as when only 100 trajectories are used for finetuning.  
In the \textit{FrozenLake} task, the value function assigns the best value to the action \textit{move right}. Although the right cell appears closer to the goal, it is actually surrounded by several traps, leaving no path to the goal.  
In the \textit{Mini-BEHAVIOR} task, the value function favors \textit{turn right} since it wrongly thinks it can go left straight to pick up the object. However that way is blocked by the table.
In the \textit{Language Table} task, the value function thinks the task is finished, but the \textit{green star} is not placed in the right position (\textit{top center}) yet.

\subsection{Qualitative Analysis of VLMs}
\label{appendix:vlm_analysis}
\begin{figure}[t]
    \vspace{-8mm}
    \centering
    \includegraphics[width=\linewidth]{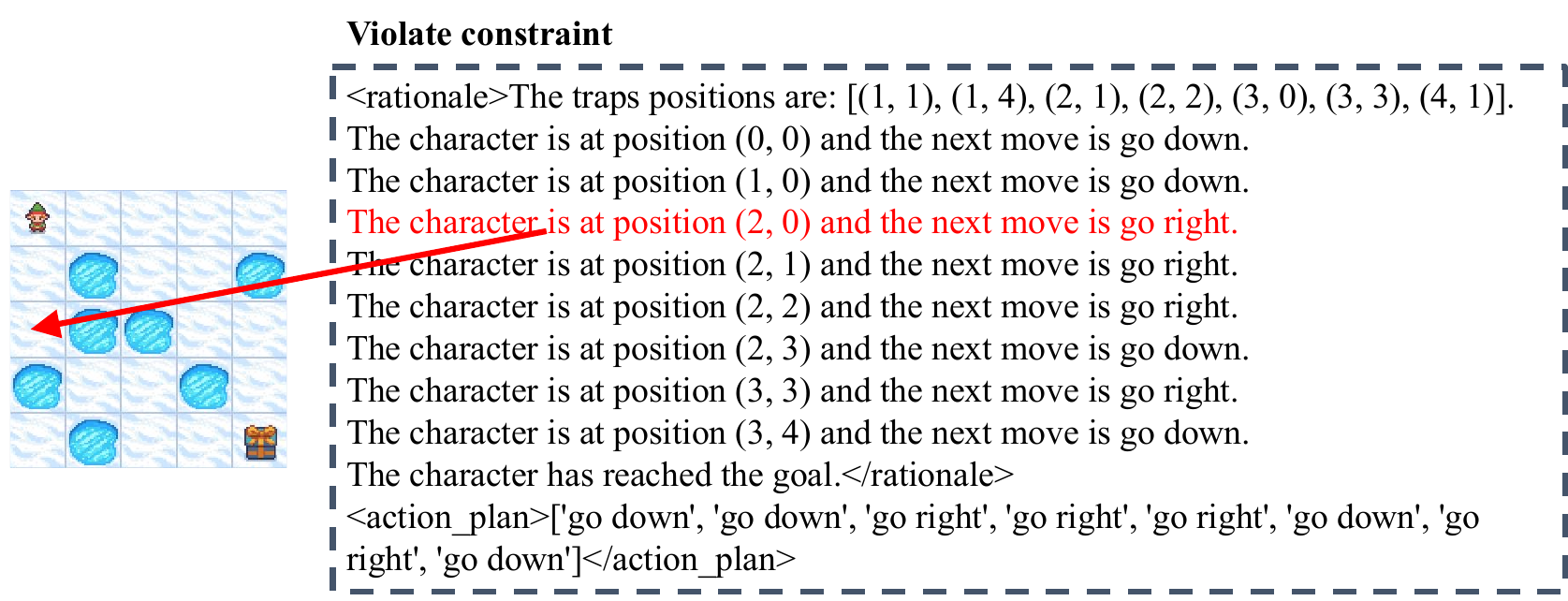}

    \vspace{1mm}\rule{\linewidth}{0.4pt}\vspace{2mm}

    \includegraphics[width=\linewidth]{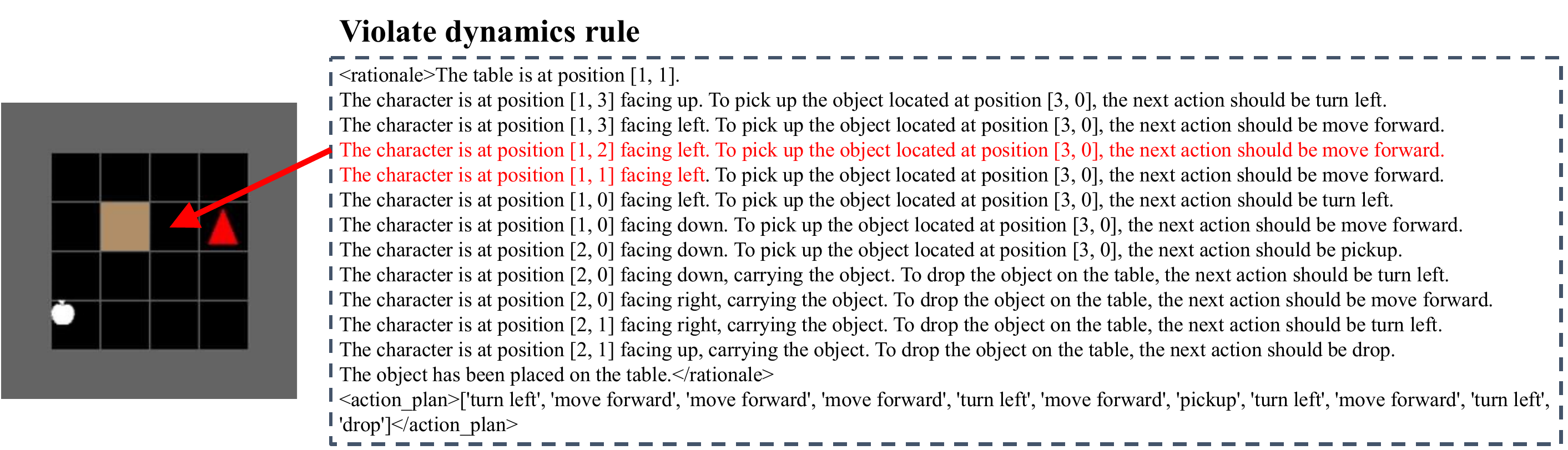}

    \vspace{1mm}\rule{\linewidth}{0.4pt}\vspace{2mm}

    \includegraphics[width=\linewidth]{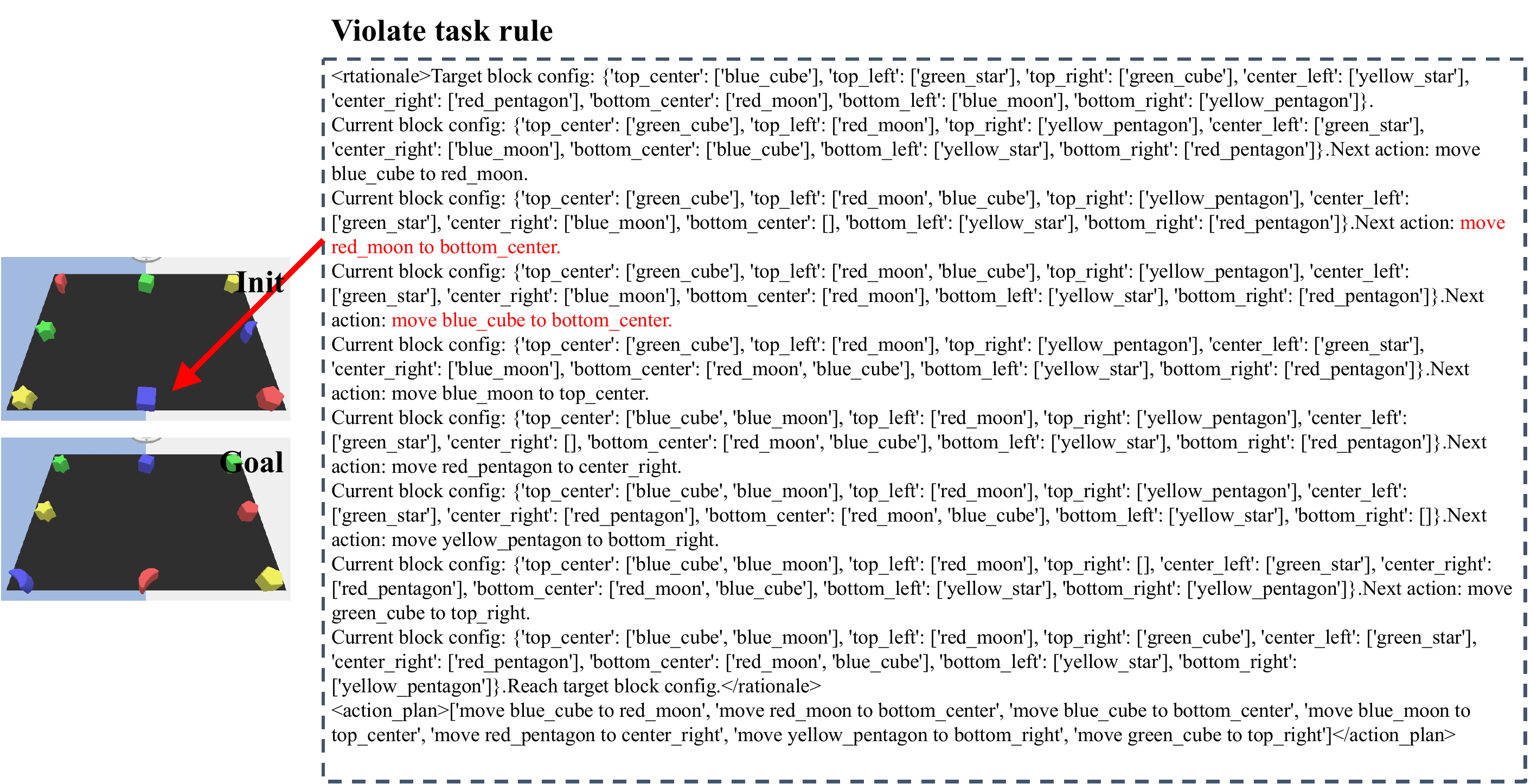}

    \caption{Qualitative analysis of errors of CoT-based planning with Qwen2.5-VL-32B-Ins-CoT on FrozenLake, Mini-BEHAVIOR, and Language Table.}
    \label{fig:qwen_cot_errors}
\end{figure}

To highlight the limitations of using VLMs for planning, we present several failure cases of Qwen2.5-VL-32B-Ins-CoT in Figure~\ref{fig:qwen_cot_errors}.

We identify three distinct factors contributing to these failures. In the \textit{FrozenLake} task, the model violates the safety constraint: it recognizes a trap at position (2,1) but still chooses to move right from (2,0). In the \textit{Mini-BEHAVIOR} task, the plan disregards environment dynamics, attempting to move forward when the agent is at (1,2) despite a table blocking the path. In the \textit{Language Table} task, the plan breaks the task rule that only one object may occupy a given position. By the third step, the bottom-center cell is already occupied by another block, yet the model attempts to move the blue cube there.
We argue that these errors stem from the fact that chain-of-thought reasoning does not explicitly construct a world model to predict the outcomes of different actions or use a reward mechanism to evaluate those outcomes. a limitation also noted in prior work~\citep{rap,gpt-mcts}.

\section{Omitted Experiments}



\subsection{Implementation with other UMMs}
\label{appendix:other_umms}
To examine whether Uni-Plan generalizes beyond BAGEL, we further implement the same framework with another recent UMM backbone, Lumina-DiMOO~\citep{lumina_dimoo}. We compare its direct VLM-style usage (\textit{VLM \& VLM-CoT}) against the full Uni-Plan instantiation on four tasks. As shown in Table~\ref{tab:lumina_dimoo}, equipping Lumina-DiMOO with our planning framework consistently yields large gains across all tasks. These results provide further evidence that the effectiveness of Uni-Plan does not depend on a specific UMM implementation, but instead transfers to other strong unified multimodal backbones.

\begin{table}[h]
\centering
\small
\caption{Results of implementing Uni-Plan with Lumina-DiMOO.}
\label{tab:lumina_dimoo}
\resizebox{0.9\textwidth}{!}{
\begin{tabular}{lcccc}
\toprule
\textbf{Method} & \textbf{FrozenLake} & \textbf{Mini-BEHAVIOR} & \textbf{Language Table} & \textbf{Real World} \\
\midrule
Lumina-DiMOO (VLM) & 0.34 & 0.18 & 0.26 & 0.25 \\
Lumina-DiMOO (VLM-CoT) & 0.44 & 0.50 & 0.56 & 0.30 \\
Lumina-DiMOO (Uni-Plan) & 0.90 & 0.86 & 0.78 & 0.70 \\
\bottomrule
\end{tabular}
} 
\end{table}

\subsection{Comparison to planning approaches with LLMs}
Besides comparing different VLMs, we also include a classic LLM-based planning baseline, SayCan~\citep{saycan}. SayCan requires iterating over the entire action (skill) space, making it infeasible to evaluate on the Language Table task. Therefore, we omit this setting. The core component of SayCan is an affordance function that determines which actions are feasible in the current state. For FrozenLake, all actions (left / down / right / up) are always valid and thus require no affordance filtering. For Mini-BEHAVIOR, “pick up’’ is feasible only when the object is not currently held, and “drop’’ is feasible only when an object is held; we use ground-truth simulator information to implement this affordance logic. For the real-world rearrangement task, we design a small rule-based affordance module to follow the SayCan procedure.

\begin{table}[h]
\centering
\small
\caption{Success rates of SayCan and our method.}
\label{table:saycan}
\resizebox{0.7\textwidth}{!}{
\begin{tabular}{lccc}
\toprule
\textbf{Method} & \textbf{Frozen Lake}
& \textbf{Mini-BEHAVIOR}
& \textbf{Real World} \\
\midrule
SayCan                 & 0.32 & 0.22 & 0.40 \\
Uni-Plan        & 0.95 & 0.83 & 0.63 \\
\bottomrule
\end{tabular}
}
\end{table}

Table~\ref{table:saycan} compares SayCan with our method. The key difference is that SayCan relies on an explicit affordance module to score the feasibility of each candidate skill in the current state, whereas our method instead performs model-based planning with a learned dynamics model and value-guided search. Despite this difference in formulation, Uni-Plan achieves substantially better performance across all evaluated tasks, indicating that planning with imagined future trajectories provides a stronger decision-making mechanism than selecting actions solely based on immediate affordance estimates.


\end{document}